# Computational Experiments Meet Large Language Model Based Agents: A Survey and Perspective

Qun Ma, Xiao Xue, Deyu Zhou, Xiangning Yu, Donghua Liu, Xuwen Zhang, Zihan Zhao, Yifan Shen, Peilin Ji, Juanjuan Li, Gang Wang, Wanpeng Ma

***Abstract*—Computational experiments have emerged as a valuable method for studying complex systems, involving the algorithmization of counterfactuals. However, accurately representing real social systems in Agent-based Modeling (ABM) is challenging due to the diverse and intricate characteristics of humans, including bounded rationality and heterogeneity. To address this limitation, the integration of Large Language Models (LLMs) has been proposed, enabling agents to possess anthropomorphic abilities such as complex reasoning and autonomous learning. These agents, known as LLM-based Agent, offer the potential to enhance the anthropomorphism lacking in ABM. Nonetheless, the absence of explicit explainability in LLMs significantly hinders their application in the social sciences. Conversely, computational experiments excel in providing causal analysis of individual behaviors and complex phenomena. Thus, combining computational experiments with LLM-based Agent holds substantial research potential. This paper aims to present a comprehensive exploration of this fusion. Primarily, it outlines the historical development of agent structures and their evolution into artificial societies, emphasizing their importance in computational experiments. Then it elucidates the advantages that computational experiments and LLM-based Agents offer each other, considering the perspectives of LLM-based Agent for computational experiments and vice versa. Finally, this paper addresses the challenges and future trends in this research domain, offering guidance for subsequent related studies.**

## I. INTRODUCTION

The integration of social, physical, and information components in complex systems, facilitated by advancements in information technologies like the Internet, Internet of Things (IoT), and Big Data, has given rise to Cyber-Physical-Social Systems (CPSS) [1,2,3]. The problems of designing, analyzing, managing, controlling and integrating such complex social systems tend to become progressively more complex with the uncertainty existing in behaviors of humans and societies [4,5,6]. In order to tackle these challenges, computational experiments have emerged as a valuable tool for humans to deduce and analyze the patterns and trends exhibited by complex social systems under hypothetical circumstances [7,8,9]. As shown in Fig. 1, computational experiments offer a distinctive "generative explanation" pathway, which encompasses both generative deduction and generative experiments [10,11,12]: ***1) Generative experiments*** are used to validate certain theories or hypotheses by introducing artificial interventions into artificial societies [13], generating data that is highly suitable for investigating causal relationships. ***2) Generative deduction*** employs deductive simulation to replicate real-world settings, or construct and observe potential alternative worlds.

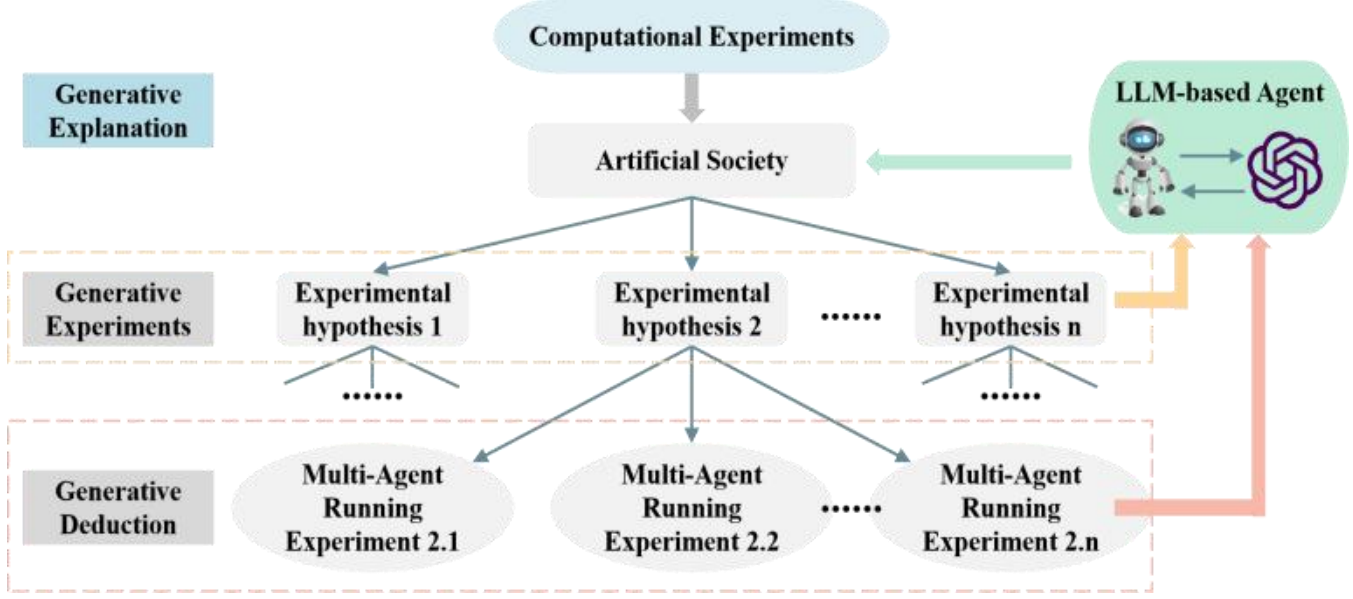

Fig. 1. Schematic diagram of computational experiment.

To enhance the credibility of computational experiments, it is crucial to ensure that the artificial society accurately reflects the complex characteristics observed in real systems. These include individual intelligence, heterogeneity, autonomy, collaboration, competition among organizations, and the emergence of complex social systems. Agent-based Modeling (ABM) [14] serves as a fundamental methodology for constructing artificial societies in computational experiments. ABM involves the creation of agents (virtual individuals) and the formulation of interaction rules to simulate real-world emergence. The typical structure of ABM comprises four modules: perception, reasoning, action, and optimization. Various techniques are integrated with ABM to achieve key agent functionalities. In order to incorporate the feature of learning evolution, machine learning can hard-code the agent's behavioral rules into an adaptive model, which improves the decision-making mechanism based on state changes [15]. To account for bounded rationality, it's necessary to integrate contemporary research in crowd psychology into the models [16]. In addition, Big Data and Social Network can facilitate the incorporation of heterogeneity and interaction among agents within ABM [17,18]. However, agents constructed using ABM possess certain limitations due to inherent model constraints and the scarcity of training data, which result in the following shortcomings:

This work has been supported in part by National Key Research and Development Program of China (No.2021YFF0900800), National Natural Science Foundation of China (No.61972276, No.62206116, No.62032016), New Liberal Arts Reform and Practice Project of National Ministry of Education (No. 2021170002), Tianjin University Talent Innovation Reward Program for Literature & Science Graduate Student (C1-2022-010), and Shanxi Province Social Science Foundation (No.2020F002).
(Corresponding author: Xiao Xue; e-mail: jzxuexiao@tju.edu.cn)

1) Lack of generality. Researchers can define a variety of structures and evolutionary mechanisms of agents according to their own understanding and the needs of the application domain, and no specific framework can be utilized to guiding the construction of uniform interaction rules between agents;
2) Lack of human-like characteristics. Agent does not have bounded rationality, so there is no cognition of the environment in which it is located; when faced with complex tasks, Agent does not have the reasoning ability, and cannot even provide relevant feedback; Agent does not have the ability to learn independently, and can only carry out the task according to the predefined rules in a step-by-step manner;
3) Lack of sociability. Due to the uncertainty of human behaviour and the complexity of the different segments in the society, it's not possible for agents to reproduce real-life behaviors perfectly in the artificial society.

These defects above lead to the fact that agents constructed by ABM are not able to comprehensively perform human's complex behaviors in real social systems, and that the artificial society is unable to represent the complexity of the social system's features. Therefore, these facts greatly increase the difficulty of using computational experiments and reduces the credibility of the method. The emergence of large language models (LLMs) [19,20,21,22,23,24] provides an opportunity for a breakthrough in computational experiments.

LLM-based Agent has demonstrated surprising capabilities in natural language processing. An increasing number of research areas use LLMs as central controllers to build agents [25,26,27,28] for key human capabilities (in-context learning, continuous learning, reasoning, etc.). Artificial societies constructed by LLM-based Agent also exhibit similar characteristics to real social systems (organized collaboration, rational competition, information dissemination, and group emergence) [164,169,211,212]. Such artificial societies are more consistent with CPSS than the artificial society constructed by ABM, and help with computational experiments on the study of complex systems. But the intrinsic mechanisms of LLM are not yet clear [29].

There are two reasons to improve the explainability of LLM. On the one hand, explainability clarifies the reasoning mechanism behind a predictive model in a way that is understandable to the user, which allows users to understand the capabilities, limitations, and potential pitfalls of LLM [30,31]. On the other hand, researchers can use explainability as a tool to rapidly improve model performance for downstream tasks and to develop reliable, ethical, and safe models for real-world deployment [32,33,34]. However, there are five difficulties in providing explainability for LLMs: **(i) High Model Complexity.** LLMs models are huge in size, containing billions of parameters, and their internal representation and inference processes are so complex that it is difficult to give explanations for their specific outputs [35]. **(ii) Data-dependency.** LLMs rely on large-scale text corpus during the training process [19], and it is difficult to judge the effect of the quality of the training data. **(iii) Black Box.** It is difficult to explicitly determine the chain of reasoning and decision-making process within the LLM, which can only be analyzed based on the inputs and outputs [36]. **(iv) Output Uncertainty.** The output of LLMs is often uncertain, i.e., it may produce different outputs for the same input [37]. **(v) Insufficient Evaluation Metrics.** The current automated assessment metrics are not sufficient to fully reflect the explainability of LLMs [181]. Therefore, these difficulties lead to lack of explicit explainability of LLMs [29], and LLM-based Agent's behaviors may lack clear causal relationships and explainability (e.g., hallucination [38,39,40]). Furthermore, it barriers to the use of LLM-based Agent in computational experiments.

In summary, this paper proposes the integration of LLM-based Agent and computational experiments to enhance the modeling of complex systems. The advantages are as follows:

1) LLM-based Agent for Computational Experiments. LLM provides Agent with anthropomorphic abilities (reasoning and autonomous learning) and new modules (memory, reflect, etc.) are embedded in agent's structure to support these abilities. Then LLM-based Agent possesses characteristics resembling humans, such as bounded rationality, heterogeneity, autonomous decision-making, learning evolution and interaction. Artificial society also exhibit some complex features present in real social systems, such as organized cooperation, game under ethical and moral principles, social communication, and social emergence. LLM-based Agent makes the whole artificial society more consistent with real-world CPSS.
2) Computational Experiments for LLM-based Agent. Computational experiments serve as a bridging tool to improve the explainability of LLM-based Agent and enhance its ability of decision-making. **(i) generative experiments.** According to various experimental designs, computational experiments can evaluate the abilities of LLM-based Agent (tool use, environment perception, etc.) and help it to clarify the causal relationship between complex social features and its output behaviors, thus enhancing its explainability [41,92]; **(ii) generative deduction.** Computational experiments can assist LLM-based Agent in simulating and deducing the potential future scenarios for each decision, which reveals the causal relationship between individual behaviors and complex social phenomena, thus achieving decision-making intelligence for LLM-based Agent.

The rest of this paper is organized as follows. In Section II, we clarify conceptual foundations of Agent, Artificial Society and Computational Experiments. Section III describes advantages of LLM-based Agent for Computational Experiments. In Section IV, we discusses advantages of Computational Experiments for LLM-based Agent. Section V presents some challenges and future research focus. Our conclusions are provided in Section VI.

## II. Conceptual Foundations: From Agent-based Modeling to Computational Experiments

This section is aimed at clarifying the relationship among Agent, Artificial Society and Computational Experiments. It begins with a review of the traditional approach to building agents, known as ABM, including employed structures and corresponding implementation techniques. Subsequently, it illustrates the combination of Agent with environmental models and rule models to compose Artificial Society. Finally, Computational Experiments are emerged to synthesize the Agent and Artificial Society.

### *A. Development of Agent Modeling: A Brief History*

Agents have a certain degree of autonomy that corresponds to individual organisms or groups of organisms in the real world. Researchers have conducted extensive research on agent modelling to study real-world behaviors in the virtual world. As shown in Fig. 2, the structure of agent has gone through four stages according to agent's intelligence [42]: reactive agent, deliberative agent, hybrid agent and adaptive agent. **(i) Reactive Agent** achieves decision-making through a "task-behaviour" collection directly mapping perceived inputs into actions. It has a simple structure that requires low constructing cost, but this type of agent can only make decisions based on local information and cannot handle situations where goals conflict with each other. Existing works include Subsumption architecture [43,44], Agent Network architecture [45], Pengi [46,47], situated automata [48], BOD [49], etc. **(ii) Deliberative Agent** has the ability to describe the environment and to reason logically. Its behaviors result from processing stages such as perception, modelling and planning. The structure of deliberative agent consistent with the Belief-Desire-Intention (BDI) structure proposed by Rao and Georgeff [50]: Belief represents the agent's perception of the environment and its own internal state; Desire denotes the agent's specific goal; and Intention symbolizes the pairs of behaviors and a plan that the agent needs to follow and adopt. This type of agent has a clear logical architecture, but is still incapable of accomplishing complex dynamic problems of representing and reasoning about the physical environment. Existing works include PRS-CL [51], Soar [52], MINDER1 [53], GRUE [54], etc. **(iii) Hybrid Agent** contains a variety of subsystems with different configurations, which are arranged in a hierarchical structure with interactions between levels. It can handle reactive and long-term planning behaviors separately, but performs not well in real-world complex reasoning problems and cannot update the decision-making mechanism in a timely manner. Existing works include Spartacus & MBA [55], EMIB [56], DIARC [57], MADbot [58], ABAsim [59], SPADE [60], etc. **(iv) Adaptive Agent** is the dominant structure currently used for artificial society modelling of computational experiments. The internal information control flow links the four modules (Perception, Decision, Reaction and Optimization) into a whole. This type of agent can collect information from the environment to generate corresponding plans, and then adjust its behaviors. After completing a round of behaviors, it can optimize the decision-making mechanism based on the feedback results. It shows a certain degree of autonomy, but still cannot reach the degree of human-like intelligence in the processing of actual complex problems. Existing works include MetaMorph [61], analytical models of epidemic spread [62], analytical models of social network communication [63], autonomous driving test [64], AdA [65], etc.

Along with the optimization of agent's structure, ABM is coupled with many new techniques to refine the implementation of four critical features, as shown in Table I. It illustrates four features that agents can exhibit, including learning evolution, bounded rationality, heterogeneity, and interaction. Furthermore, it describes how four technologies (Machine Learning, Psychological Model, Big Data and Social Networks) help agents exhibit the above four features with some existing works.

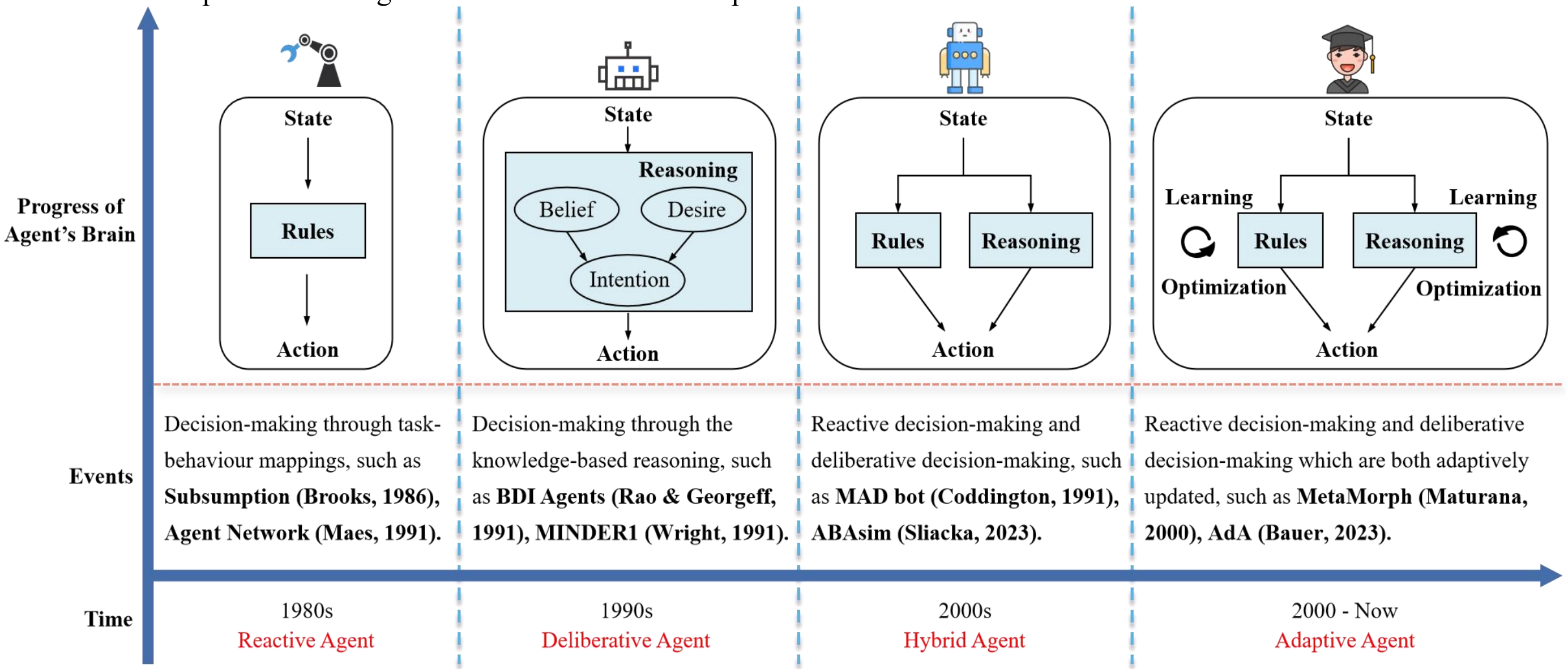

Fig. 2. The structure development of agent (reactive agent, deliberative agent, hybrid agent, adaptive agent).

TABLE I
TECHNOLOGIES COUPLED TO FOUR TYPES OF CRITICAL FEATURES IN ABM

| Features | Explanation | Technology |
|---|---|---|
| **Learning and Evolution** | Agents need to have the capability to gain experience from perceived information to modify or enhance their priori knowledge. Machine learning is used to achieve learning and evolution in ABM. | A machine learning-based inference model can hard-code the agent's behavioral rules into an adaptive model that improves the decision-making mechanism based on state changes [66]. (i) Supervised learning methods are used for situational awareness learning of environment [67], agent behavioral prediction [68], agent behavioral interventions [69] and macro emergence analysis [70]. (ii) Unsupervised learning methods can be used to detect the patterns of the macro-level emergence of ABMs that cannot be easily recognized by human experts and specialists [71]. (iii) Semi-supervised learning can also be used for situational awareness learning to help ABMs learn induction rules [72]. (iv) Reinforcement learning can learn agents' behavioral distribution or strategy distribution, thus enabling the prediction of the agents' behaviour [73] as well as behavioral interventions [74]. Furthermore, RL algorithms can be used for the reinforced decision-making of macro-agents [75]. |
| **Bounded Rationality** | Agents have limited knowledge and capabilities, which leads to the fact that they may not be able to make decisions that maximize utility. Contemporary research in crowd psychology [76] is embedded in ABM for bounded rationality. | One economist, Thomas Schelling, explored human segregation through the ABM [77]. Kalick and Hamilton implement agent simulations based on the psychology of opposite-sex attraction [78]. Lorenz et al. [79] developed a mathematical model of attitude change that integrates diverse theories in social psychology in a coherent manner, and thus makes them applicable in agent-based models. Castro et al. [80] used ABM to simulate human's risky decision-making in financial markets, and compared the accuracy of two artificial markets based on Expected Utility Theory and Prospect Theory. |
| **Heterogeneity** | There need to be differences between agents, such as attribute differences and behavior differences. Big data helps ABM analyze human characteristics and behaviour for heterogeneity. | Big data analysis gather useful insights from large-scale and real-time data sources to develop an understanding of real-world phenomena [81]. Big data can help ABMs to not only identify demographic characteristics and personality attributes of the population (hometown, gender, ethnicity and population distribution) [82,83] but also understand human mobility and identify movement patterns of people [84,85]. |
| **Interaction** | Sensible interaction rules need to be defined between agents. Social networks can describe relationships among agents and agents groups to achieve interaction in ABM. | Social networks help ABMs understand various social phenomena [86]. Social networks can be used for a variety of communications within and between agent groups [87,88]. In addition, agent's position in the social network generates social capital for itself, which in turn generates behaviors such as self-organization [89], trust [90], support [91], etc. |

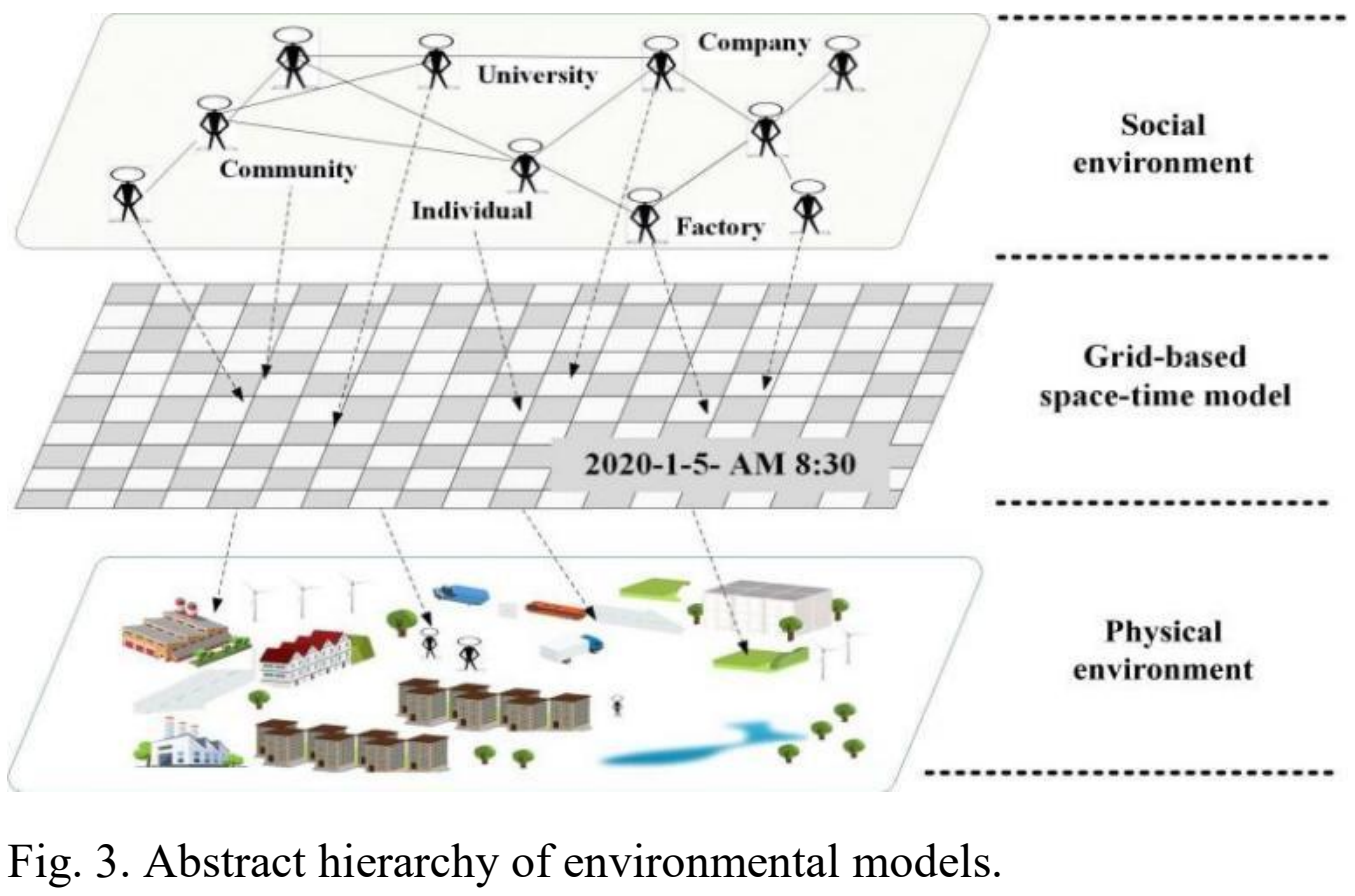

Fig. 3. Abstract hierarchy of environmental models.

*B. From Agent to Agent Society*

The most important step in modelling of artificial society is to abstract various types of models from real complex systems, including individual model, environmental model and social model [92]. ABM can help computational experiments to construct individual model, then the next step is to combine with environmental model and social model.

In Artificial Society, environmental model is a mapping of the actual physical environment in the computer, including physical features and social features, as shown in Fig. 3. Physical features represent external elements of a real society such as buildings, road traffic, climatic conditions [102], and internal attributes [96] such as the type of society, carrying capacity, relative location, etc. Social features include statistical characteristics of agents (the total number of agents, gender ratio, age distribution, etc.), geographical distribution of agents, and social relationships. Computational experiments can utilize grid-based modelling to achieve the representation of physical features. For social features, the idea is to recover the statistical features from the group data to reconstruct the specific features of each individual in the group. This process needs to satisfy the consistency of the generation law of agents with that of the real world [93] and the consistency of the internal logical structure and association relationship of the generated agents with those of the real world.The circular mechanisms of Artificial Society is defined by rule model, including guidelines for interaction among agents, environments, Agent and Environment. Social learning evolution (SLE) [94,95] formulates the evolutionary process of Artificial Society at three layers: The bottom layer is the individual evolution space, which is used to simulate the phenomenon of genetic evolution in social systems; the middle layer is the organization evolution space where agents enhance their ability through observation and imitation; The top layer is the space of social evolution, which simulates the emergence of social culture that draws best knowledge from the bottom layer and guides the evolution of individuals at the bottom layer.

Based on the research targets, the models of artificial society constructed by ABM can be classified into three categories: thought experiment, mechanism exploration, and parallel optimization, as shown in Table II.

TABLE II
APPLICATION CASES OF ARTIFICIAL SOCIETY

| Categories | Explanation | Examples |
|---|---|---|
| **Thought Experiment** | Thought experiment does not model a specific scenario or specific real social system but pursues the abstract logical relationships that describe general social systems. It is hoped to explore and quantitatively analyze the unpredictable results of certain hypotheses on human society through experiments. Agents in such artificial societies do not have the ability to learn so that they can only make decisions and interact according to behavioral rules predefined by the researcher. Examples include SugarScape Model [96], Schelling Model [77], RebeLand Model [97], and Landscape Model [98]. | **SugarScape Model** proposed by Epstein et al. [96] allows for related experiments in economics and other social sciences, as shown in Fig. 4(a). A simple set of agents moves around a grid in search of "sugar" (a food resource that is abundant in some places and scarce in others). In spite of its simplicity, this model gives rise to surprisingly complex group behaviour (migration, fights and isolation in the neighbourhood). Ultimately, only a few agents have a lot of sugar, while the majority have little sugar, demonstrating the famous Matthew effect in sociology [99]. If SugarScape were to have several additional resources, it would be possible to study the formation of markets through the exchange of resources between individuals in real societies. |
| **Mechanism Exploration** | Mechanism exploration is a modeling of a real social system, with emphasis on high matching between artificial society and real social systems. Agents in such artificial societies have a certain evolutionary capacity. They are bounded rational beings capable of learning and adapting to their environment. They are able to optimize their decision-making mechanism based on the results of past predictive feedback to decide whether to cooperate or compete with other agents. Examples include Artificial Stock Market [100], Service Bridge Model [101], and epidemic spread models (EpiSimS [102], CovidSim [103], GABM [104]). | Peng [100] proposed an **Artificial Stock Market** based on the real situation of China, through the method of heterogeneous agent-based modeling with bounded rationality, as shown in Fig. 4(b). A set of trading agents make predictions by observing changing stock prices and dividends in the digital world. Based on these predictions, they decide whether and how much to buy in order to maximize their own utility. In turn, the decisions of all traders determine the state of competition between supply and demand. |
| **Parallel Optimization** | Parallel optimization, through the establishment of an artificial social model that has a homomorphic relationship with the real society, realizes the parallel execution and cyclic feedback between these two, thereby supporting the management and control of the real complex system. Examples include The AI Economist [105], Virtual Taobao [106], Autonomous Driving [107,108]. | Fig. 4(c) shows AI Economist based on **Two-layer Reinforcement Learning** [105]. In this thesis, economic simulations are carried out by means of agent-based modelling in order to discover tax strategies that efficiently strike a balance between economic equality and productivity. Citizens can learn in response to economic goals and thus optimize their behaviors as well as tax strategies. |

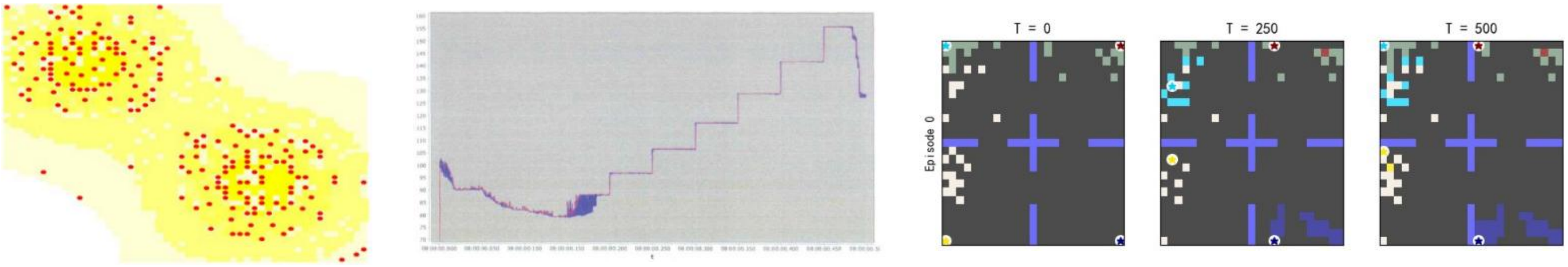

(a) Sugarscape (b) Artificial stock market (Peng [100]) (c) AI Economist

Fig. 4. Schemas of Artificial Society. (a) Distribution of sugar and Agent in Sugarscape. (b) The price curve in the game between institutional investors and retail investors. (c) Distribution of AI Economists and all types of resources in the society.

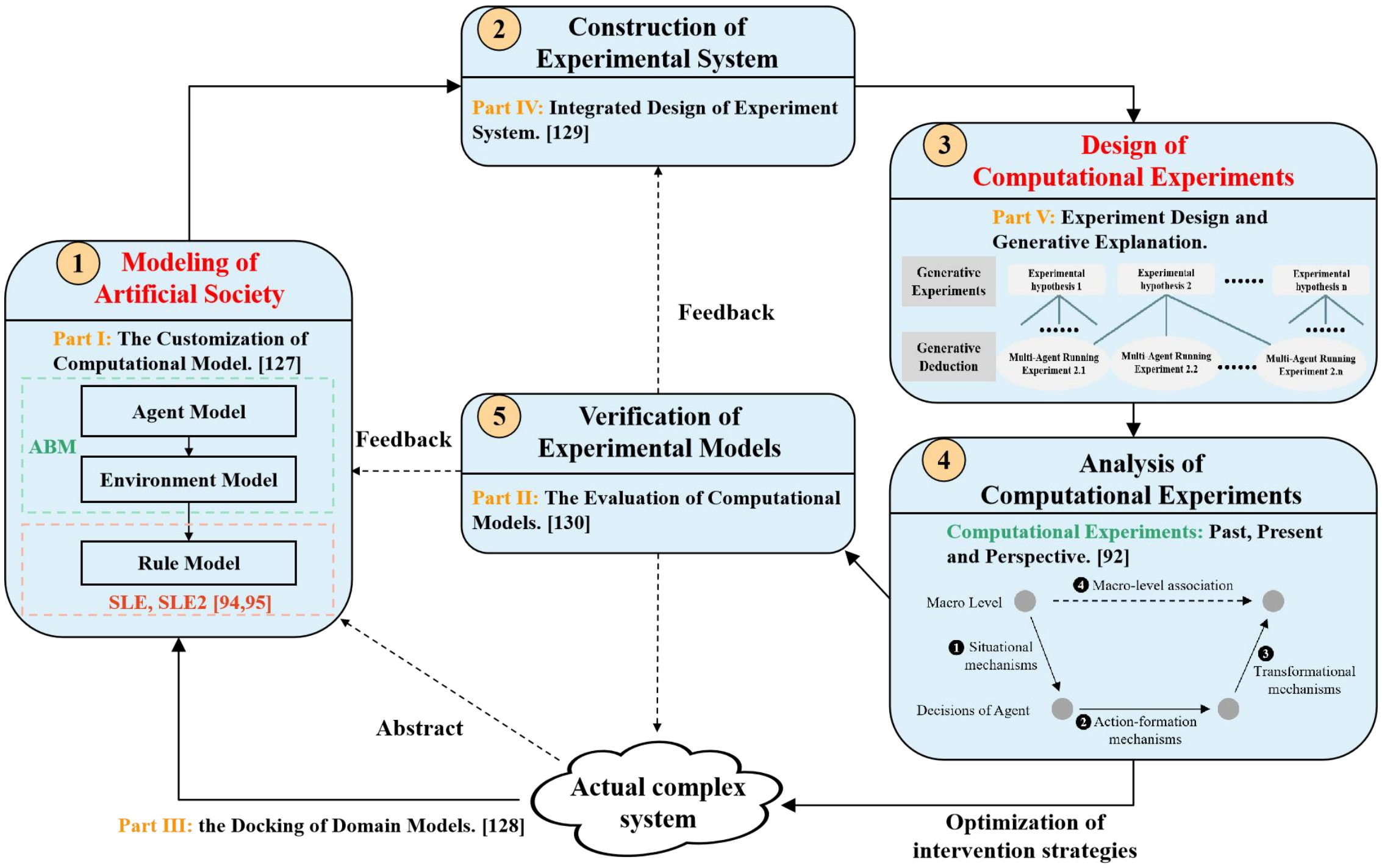

Fig. 5. Methodological framework for computational experiments.

### *C. The Emergence of Computational Experiments*

With the progress of ABM technology and the rapid development of artificial society modeling, the research on complex systems has also made great progress, paying more attention to the adaptability and dynamics of the system [109,110,111]. Nowadays, the emergence of CPSS has led the study of complex systems into a new era [112,113,114], prompting computer simulation techniques, such as ABM and artificial society, to be combined with emerging data science theories to explore the intrinsic laws of complex systems. As a result, computational experiments have emerged and the ACP approach (Artificial Societies + Computational Experiments + Parallel Execution) [115] has been developed, which emphasizes the cyclic feedback-optimization relationship between artificial systems and real systems [116,117,118].

Computational experiments utilize computers as "artificial laboratories" [119,120,121] to "incubate" macroscopic phenomena that may occur in real systems, and to explore the laws behind social emergence [122,123,124]. Computational experiments provide a viable way to analyse the behaviour of complex systems and to evaluate various intervention strategies [125,126]. The methodological framework consists of five steps: modeling of artificial society [94,95,127,128], construction of experimental system [129], design of computational experiments [12], analysis of computational experiments [92], and verification of computational experiments [130], and forms a feedback loop shown in Fig. 5.

Modeling of artificial society is the first step of computational experiments, and the artificial society is the basis for performing the subsequent steps. Agents in the artificial society have knowledge and learning mechanisms similar to those of real individuals, and the evolutionary mechanism of the artificial society is consistent with that of the real world. This allows artificial societies to simulate and deduce future social scenarios, i.e. generative deduction, to provide decision support for the research of complex systems [131]. In addition, computational experiments can conduct a wide variety of sociological experiments in artificial societies [132] by modifying agents' behavioral rules and various types of parameters, i.e. generative experiments, and thus explain and understand macroscopic phenomena in real social systems.

Most artificial societies have been implemented by researchers using Agent-based Modeling (ABM) [14] as a basic methodology. ABM can reproduce the interactive behaviors and communication processes of heterogeneous subjects, which in turn helps researchers to analyse the generative problem: how the distributed local interactions of heterogeneous subjects can generate macroscopic rules. This explanation path is called generative explanation [133]. However, there are some unavoidable defects [134,135] in the existing Agent and Artificial Society, which make the current generative explanation for macroscopic phenomena lack of rationality and credibility, thus hindering the development of computational experiments. The defects are as follows:

#### 1) Generality

Due to the increasing complexity of social interactions in the real world, the behaviour of others, one's own memory, learning ability and emotions may lead to different human behaviors in different situations. However, there exists no unified model structure and mathematical rules to build Agent or Multi-Agent systems. For example, most researchers embed different conceptual cognitive frameworks (PECS [136], BDI [50], etc.) into the structure of Agent for modelling human behaviour [137,138], while some others lead the construction of Agent from the perspective of pattern recognition using different optimization algorithms (neural networks [139], genetic algorithms [140], etc.), focusing on the performance of Agent in various experimental tests [141,142]. These cases show that researchers define a variety of structures, evolutionary mechanisms, and interaction rules between agents according to the research objectives (simulation, optimization, evaluation), different service scenarios (industrial, educational), and their own knowledge of human behaviour. This also leads to the fact that agents and the corresponding artificial societies constructed by the researchers are highly subjective, with too much freedom and no generality in the models, thus failing to perfectly map the complex features of real social systems. As a result, it also greatly increases the difficulty of using computational experiments and reduces the credibility of the method.

**2) Human-like Characteristics**

Computational experiments have been successfully applied to studies of systems [143,144,145] that are risky, costly, or where direct experimentation is not realistically possible. As the main subject of Artificial Society, computational experiments expect Agent to have the following five types of anthropomorphic characteristics:

a) *Bounded rationality.* In real life, people's information, knowledge and ability are limited as well as the options they can consider, so that people may not be able to make decisions that maximize utility. This means that human beings need to have bounded rationality [146]. Therefore, bounded rationality is a human-like characteristic that must be considered in the process of constructing Agent.
b) *Heterogeneity.* There must be differences between agents, including attribute heterogeneity (own preferences, natural abilities, etc.) and behavioral heterogeneity [147,148]. It greatly improves the utility of agents and artificial societies. More importantly, it can help to simulate a dynamic game process with multiple subjects and multiple expectations to study the dynamic game problems that always exist in complex social systems [149].
c) *Autonomous decision-making.* Agents need the ability to make autonomous decisions based on perceived information, memory, prior knowledge and their own state. Humans with advanced intelligence can perceive the world, acquire information, reason to make decisions, create value, self-reflect and learn to evolve in the real world. If Agent only relies on the priori knowledge provided by the designer for mechanical behavioral responses, then the agent lacks autonomy [150] and cannot be used as a human substitute in an artificial society.
d) *Interaction.* The interaction between agents is necessary for the emergence mechanism of the system [151], including three categories: the interaction between agents and the environment [152]; the interaction between agents [153], including the exchange of information, mutual imitation, etc.; and the interaction between agents and tools [154,155], including API calls, obtaining additional resources, etc.
e) *Learning evolution.* Under the premise of bounded rationality, Agent has some prior knowledge as well as the ability to gather information, and it is able to gain experience based on what it perceives and modify or extend the prior knowledge.The learning model of Agent is classified into three categories: Non-conscious learning, which means that agents adjust their response to external stimuli instinctively, avoiding harm, such as reinforcement learning [156]; Routine-based learning describes that agents went through some sort of introspection about the experience and then adjusted the behaviour, such as learning direction theory [157]; Belief learning means that agents understand the world sufficiently and learn knowledge by iterating the understanding, such as neural network [158].

**3) Sociability**

Due to the complexity of human behaviour, real social systems often exhibit social behaviors that are difficult to define [159], such as cooperation, competition, subordination, etc. ABMs typically use predefined rules, such as heuristic rules or learning algorithms, to force agents to work together while researchers can't distinguish if it is cooperation or subordination. Therefore, this can lead to oversimplification of the agent, which in turn limits the accuracy of the model and greatly reduce the ability of the artificial society to reflect complex phenomena in the real world. Computational experiments expect Artificial Society to have the following four types of complex features:

a) *Collaboration.* Agents and groups can regulate their own behaviour to achieve common goals through information sharing and organizational scheduling.
b) *Competition.* Agents or groups adjust their behaviour against each other in order to obtain more social resources for themselves.
c) *Communication.* Both physical matter (food, devices, etc.) and abstract concept (information, culture, emotion, etc.) can be transmitted through social relations among agents and groups.
d) *Emergence.* Spontaneous interactions between agents or groups can create social phenomena that is complex and possibly unexpected.

## III. LLM-based Agent for Computational Experiments

With the arrival of the era of Large Language Model, Agent has gradually entered the vision of a wide range of researchers, which will serve as a carrier of LLM and continue to bear the heavy responsibility of exploring Decentralized Autonomous Organizations (DAOs) [160] and Artificial General Intelligence (AGI) [161]. LLM-based Agent demonstrates powerful problem-solving capabilities and a high degree of intelligence since LLM is utilized as its brain. This provides a breakthrough opportunity for computational experiments that need to enable the study of complex systems by modelling micro-individual behaviour and macro-social behavior.

*A. Agent Empowered by LLM*

The history of research on Agent is actually as long as the research on Artificial Intelligence. In recent years, with the rise of LLM, the topic of agent research has been brought back into the public domain, and there are a number of LLM-based agent projects, including AutoGPT [162], BabyAGI [163], Generative Agents [164], MetaGPT [154], and others [165,166,167,168,169,170,171].

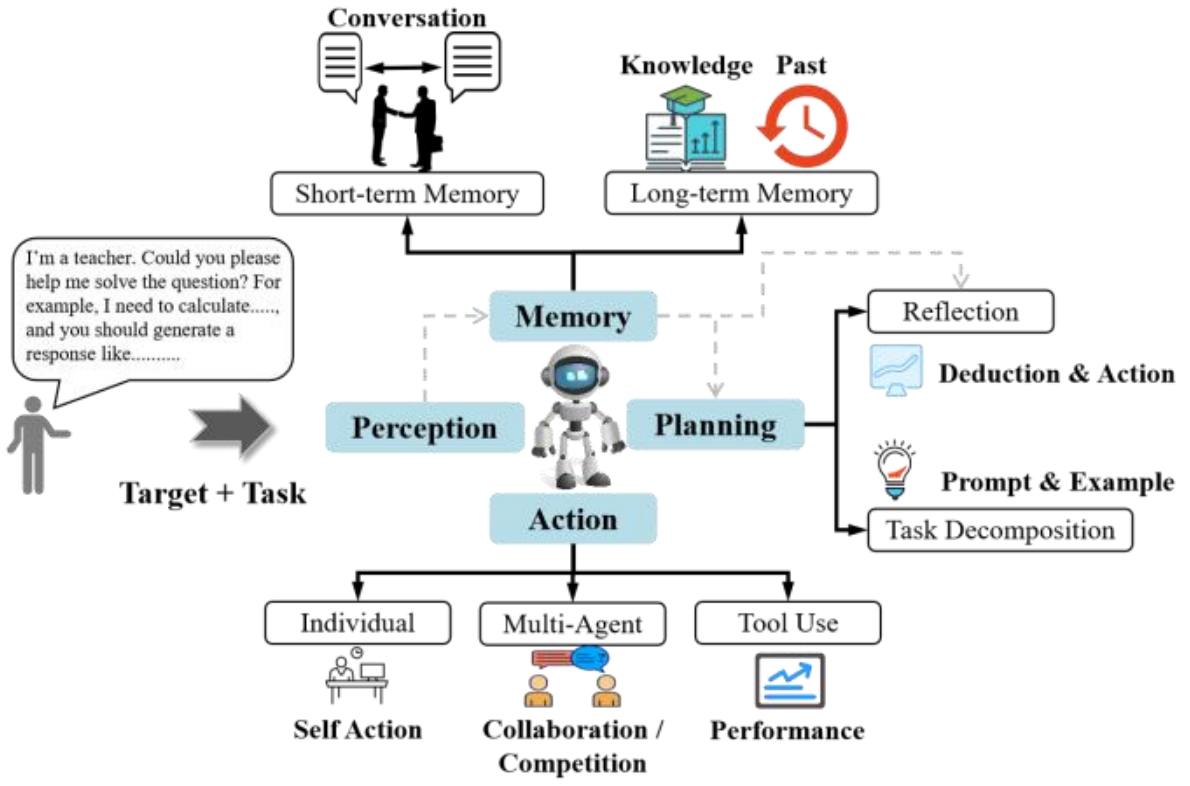

Fig. 6. The structure of LLM-based Agent.

As shown in Fig. 6, this section summarizes the common structure of LLM-based Agent, which includes four parts: **(i) Perception.** LLM-based Agent already has the basic ability to communicate with humans through textual input and output [154,162]. And it can understand the deeper meanings (desires, intentions, etc.) behind natural language [172]. **(ii) Memory.** Memory is defined as the process by which information is acquired, stored, retained and subsequently retrieved [164]. Short-term memory stores necessary information to perform complex cognitive tasks (e.g. learning and reasoning) and is equivalent to an agent's working memory; long-term memory can store information for extended periods of time and is equivalent to an agent's knowledge base [173]. **(iii) Planning.** A complex task usually consists of several sub-steps, and LLM-based Agent needs to divide the complex problem into small steps that can be solved step-by-step in advance [174]; the agent can achieve self-reflection by improving past action decisions and correcting past mistakes [175,176]. **(iv) Action.** Most of current action modules of LLM-based Agent exploits text dialogues instead, which can generate high-quality and diversified textual responses for humans [214,219]; LLM-based Agent can call external tools to significantly extend its functionality [177,178].

Compared with the traditional structure of Agent, the biggest change in the structure of LLM-based Agent is that the decision-making module and optimization module in the traditional structure are integrated into the Brain module in the new structure: LLM acts as the main body of the Planning module, and Memory module is added to help LLM to make rational decisions. The advantage of LLM over the four types of technologies (shown in Table I) is that LLM provides a variety of new capabilities for agents: reasoning, autonomous learning, role-playing, and natural language understanding, as shown in Table III.

LLM shows strong language comprehension and generation capabilities [23,179], so that LLM-based Agent can better simulate a role in the real society, which communicate with humans easily, and help humans solve complex problems. However, ABM has insufficient description of individual heterogeneity, which makes it impossible for Agent to simulate a role as well as LLM-based Agent, and lacks fidelity to human behavior [180]. This also leads to the fact that the Artificial Society constructed by ABM cannot reflect the complex phenomena of real society, and computational experiments have not made further breakthroughs in the research of complex systems.

TABLE III
COMPARISON BETWEEN ABM AND LLM-BASED AGENT

| Features | ABM | LLM-based Agent |
|---|---|---|
| **Heterogeneity** | Big data can help ABM to identify different character attributes, including personality, behavioral patterns, etc., but this can lead to difficulties in updating existing agent models, greatly increasing the complexity of ABM. | LLM provides the ability of **Role-playing** to achieve heterogeneity of LLM-based Agent [181]. Researchers can design perfect role prompts (including name, hobby, daily life, behavioral rules, etc.) according to the needs of the scenario and provide these prompts to LLM-based Agent to play the role required by researchers [164,214]. |
| **Bounded Rationality** | Psychological models are embedded in ABM to represent the bounded rationality of human beings. However, human's psychological characteristics are complex and there are limitations in using a single model to construct agents. | LLMs are pre-trained on large amounts of human data, enabling them to have **Extensive Knowledge** across multiple domains and exceeding the capabilities of general human. However, too much knowledge can undermine the bounded rationality of LLM-based Agent and reduce the credibility of the model. LLM-based Agent driven by prompting may express knowledge that is inconsistent with the character's identity and era, i.e. hallucinate [39]. A common solution is to collect character-specific events to form a dataset and fine-tune LLM so that LLM-based Agent can specify the character's abilities [182,183]. When confronted with questions beyond the limits of a character's inherent abilities, LLM-based Agent may avoid providing answers and exhibit ignorance, i.e. bounded rationality. |
| **Autonomous Decision-making** | Machine learning algorithms can provide decision-making mechanisms for agents based on existing data, but such decision-making mechanisms can often only handle simple inference problems. | **Reasoning** is an important feature to demonstrate that LLM-based Agent has human-like intelligence. For human beings, reasoning is fundamental to intelligence, serving as the cornerstone for problem-solving, decision-making, and critical analysis [184,185]. For LLM-based Agent, reasoning is the key to achieve autonomous decision-making. A lot of works [186] have shown that it can be processed with fine tuning or with appropriate bootstrapping means (CoT [187], ToT [188], etc.) to obtain a certain level of reasoning ability [189]. For example, when using CoT prompts, LLMs are able to produce valid individual proof steps, even when the synthetic ontology is fictional or counterfactual [190]; according to , LLMs exhibit reasoning patterns that are similar to those of humans as described in the cognitive literature [191]. |
| **Learning Evolution** | Machine learning algorithms are also able to optimize decision-making mechanisms based on perceived data when agents are running, but this usually requires explicit objective functions and reward mechanisms. | LLM provides agents with the ability to learn autonomously. LLM-based Agent is able to maintain the awareness of learning in interaction and continuously optimize its behaviour and decision-making mechanisms to better adapt to various tasks and environments. **(i) In-context Learning.** With the scaling of model size and corpus size [192,193], LLMs can learn from a few examples in the context. Through in-context learning, LLM-based Agent can easily acquire knowledge from natural language examples given by humans [21,194,195] and does not require any additional training process [196]; the key idea of in-context learning is to learn from analogy so that LLM-based Agent can make decisions by learning by analogy [197]. **(ii) Continuous Learning.** LLM-based Agent is capable of incrementally acquiring, updating, accumulating, and exploiting knowledge throughout its lifetime [198]. Voyager [166] can synthesize complex skills by composing simpler programs, which compounds its capabilities rapidly over time and alleviates catastrophic forgetting in other continuous learning methods. |
| **Interaction** | Social networks help agents to specify the relationship between different items, but specific interactions can only be realized through the passing of state vectors. | LLM-based Agent has the powerful language comprehension ability and generation skill, and can use **Natural Language** to interact, including collaborative development of software [155], language confrontation [219] and so on. In addition, LLM-based Agent can utilize external tools (retrieval, mathematical computing, translation, etc.) to significantly expand its functions [199,200]. |

### *B. Enhancement in Artificial Society*

For a long time, sociologists often conduct social experiments to observe specific social phenomena in a controlled environment [201,202]. However, these experiments always use organisms as participants, making it difficult to carry out various interventions and inefficient in terms of time. Therefore, an interactive artificial society is needed, in which human behavior can be performed by trusted agents [203]. From sandbox games like "The Sims" to the meta-universe, simulated society is defined as: environment and interactive individuals. Behind each individual can be a program, a real human, or a LLM-based Agent, and sociability is born from the interaction between individuals.

Artificial Society constructed by LLM-based Agent is an open, persistent, situated, and organized framework where LLM-based agents interact with each other in a defined environment [203]. Compared with the traditional artificial society, this kind of artificial society can emerge a variety of complex social phenomena, including organized collaboration, competitive game under ethical and moral principles, social communication, and social emergency, as shown in Table IV.

As a research method for complex systems, computational experiments hope that Artificial Society can perfectly map complex behaviors at all levels of real social systems. Artificial Society constructed by LLM-based Agent is increasingly capable of simulating human-like responses and behaviors, offering opportunities to test theories and hypotheses about human behavior at great scale and speed. Pretrained on massive datasets, LLMs can represent a vast array of human experiences and perspectives, giving them a higher degree of freedom to generate diverse responses than that of conventional human participant methods. LLM-based agent can also generate responses across a wider range of parameters than human participants because of pragmatic concerns of limited attention span, response bias, or habituation among humans, providing a less biased view of underlying latent dimensions [204]. So computational experiments are allowed for the testing of interventions in the simulated society (constructed by LLM-based agent) before real-world implementation.

TABLE IV
COMPARISON BETWEEN TRADITIONAL ARTIFICIAL SOCIETY AND THAT CONSTRUCTED BY LLM-BASED AGENT

| Features | Traditional Artificial Society | New Artificial Society Constructed by LLM-based Agent |
|---|---|---|
| **Collaboration** | Agents usually perform tasks based on fixed collaboration algorithms, and there is no clear division of labor. However, the collaboration process in this kind of artificial society is chaotic and unable to cope with environmental changes [205]. | In the real world, complex tasks such as software development, consulting, and game playing might require **Organized Cooperation** among individuals to achieve better effectiveness [155,216]. Artificial society brings agents together, and relevant researches focused on exploring the potential of agents' cooperation. For example, Park et al. [164] found collaboration behaviors emerge within a group of agents. Moreover, diverse multi-agent group can effectively prevent and rectify errors through interaction, ultimately enhancing decision-making capabilities during collaborative problem-solving [217]. |
| **Game** | Agents can adjust their strategies according to changes in environmental status [206], such as the artificial stock market model [100]. However, this kind of game only takes profits as the goal, ignoring the ethics existing in the real society. | Currently, LLM-based agents are usually involved as communities in experiments concerning **Game Theory**, such as the iterated Prisoner's Dilemma or other economic scenarios like ultimatum game, dictator game, and public goods game [207], in which multiple agents are organized in a way of either competition. Meanwhile, simulated societies offer a dynamic platform for the investigation of intricate decision-making processes, encompassing decisions influenced by **Ethical and Moral Principles**. For example, some researchers explore how LLM-based agents make decisions when confronted with challenges of deceit, trust, and incomplete information [208,209,210]. Furthermore, as LLMs approach AGI, LLM-based Agent have embedded in themselves human-like preferences for fairness and regard for others. |
| **Communication** | The traditional artificial society can simulate some social communication phenomena [102,103], such as the prediction of the spread trend of epidemics, impact assessment, etc., but it lacks the simulation of the spread of abstract concepts, such as emotions, public opinions, etc. | By employing LLM-based agent in the social system, the emergence of macro-level phenomena can be observed, including the **Communication of Information, Attitudes, Emotions, Culture, and Diseases**. In [169], researchers harness the human-like capabilities of large language models (LLMs) in sensing, reasoning, and behaving, and utilize these qualities to construct the S3 system; and they simulate three pivotal aspects: emotion, attitude, and interaction behaviors. There are similar works: cultural transmission [211] and the spread of infectious disease [104]. By employing LLM-based agents to model individual behaviors, implementing various intervention strategies, and monitoring population changes over time, these simulations empower researchers to gain deeper insights into the intricate processes that underlie various social phenomena of communication. |
| **Emergence** | The traditional artificial society and the real society need to have a homomorphic relationship so that it is possible to reconstruct existing social emergence and even to deduce possible future social emergence [105,106]. However, it shows emergence phenomena that tend to be the result of deterministic function, i.e. new social emergence is mostly absent as long as the key factors are unchanged. | LLM-based Agent behaviour tends towards probabilistic outcomes with great uncertainty. This also allows new artificial societies to generate more diverse **Social Emergence**, such as some emergent social behaviors. [212] constructs a structured domain knowledge graph and guides LLMs to reason over it through prompt chain, which provides evidence-based decision-making in various stages. Social simulations can assist in making informed decisions, foreseeing potential emergence phenomena, and formulating policies that aim to maximize positive outcomes while minimizing unintended adverse effects. [213] simulate scenarios involving resource extraction, pollution, conservation efforts, and policy interventions to support sustainability investigations. |

In the process of traditional computational experiments, agents' behaviors (interaction and learning evolution, etc.) shown in the artificial society tends to be the results of mathematical deterministic model. As long as the key factors are not affected, agents' behaviors will not change significantly. But for humans in the real world, everything that happens in life is possible to touch them, and could become a key factor in human decision-making. Behaviors of LLM-based Agent tends to be probabilistic with great uncertainty. This leads to the differences among agents, meanwhile also makes the whole artificial society more consistent with realistic CPSS, which in turn promotes the development of computational experiments in the research of complex systems.

### C. Application Cases

LLM has powerful natural language understanding ability and text generation ability, and its performance in agent simulation and social simulation far exceeds ABM and the traditional artificial society. LLM-based Agent can help humans deal with complex tasks through multi-agent interaction, including multi-agent collaboration and multi-agent competition. In addition, LLM-based Agent promotes the development of social simulation and makes artificial society more consistent with CPSS.

### 1) Multi-Agent Collaboration

At present, cases of multiple LLM-based agents collaboration are very common. Their implementation ideas are basically similar: humans throw out task requirements, and agents communicate and collaborate to complete the tasks. CAMEL [214] achieves collaboration by allowing two LLM-based Agents to role-play (user and assistant); AutoGen [215] allows developers to build LLM applications by customizing agents, allowing them to talk to each other to complete tasks; AgentVerse [216] can dynamically adjust the architecture of multiple agents, and allows agent groups to combine into a system. In addition, Park et al. [164] constructed an LLM-based agent with a memory flow, called Generative Agent, and instantiate generative agents to populate an interactive sandbox environment inspired by The Sims, where end users can interact with a small town of twenty-five agents using natural language, as shown in Fig. 7. These generative agents produce believable individual and emergent social behaviors. For example, some agents can work together to organize a Valentine's Day party through spontaneous communication in a simulated town.

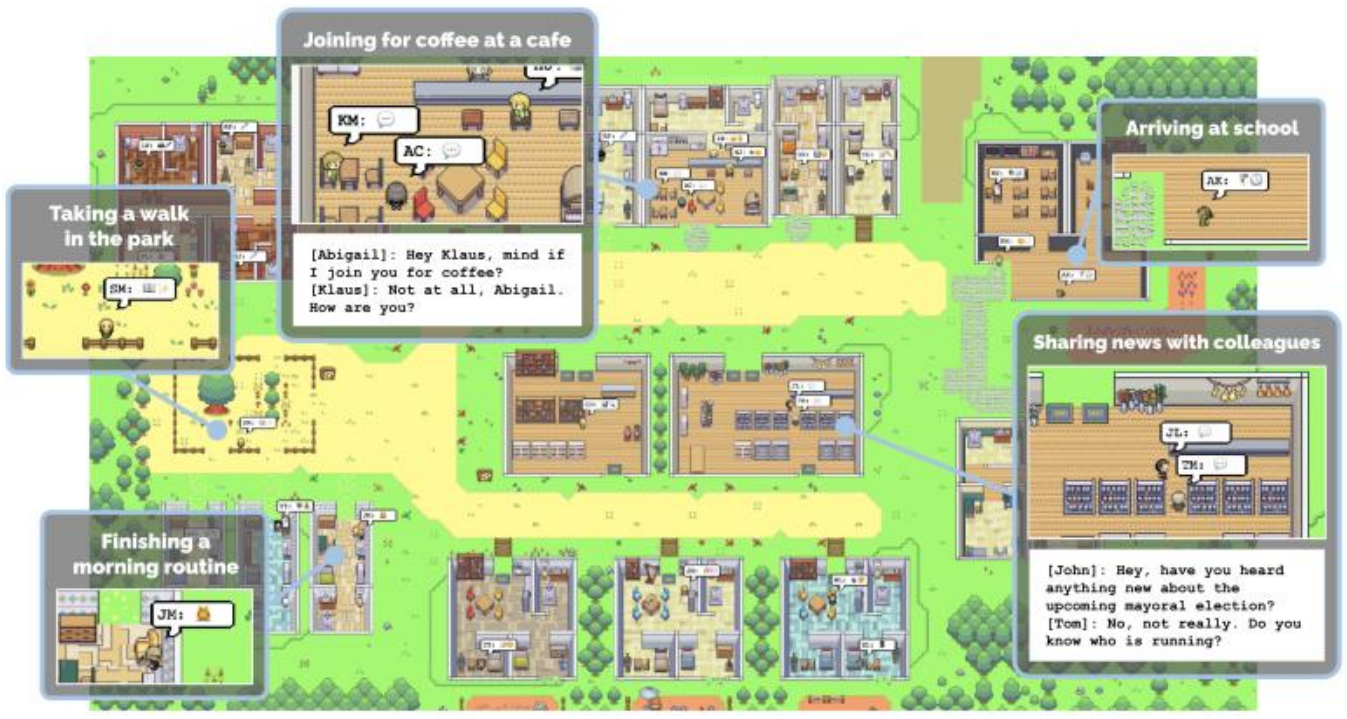

Fig. 7. Generative Agents (Park et al. [164]).

### 2) Multi-Agent Competition

Although collaboration has been widely explored and performed well in multi-agent task processing, but in real life, the emergence of competition can lead to more robust and effective human behavior. Therefore, agents can quickly adjust their strategies through interaction and strive to choose the most advantageous or rational behaviors to respond to changes caused by other agents and the environment. ChatEval [217] is a multi-agent referee team that is used to autonomously discuss and evaluate the quality of responses generated by different models, which imitates the human evaluation process; CompeteAI [218] simulates a virtual town containing restaurant agents and customer agents to study the competition between agents. In addition, researchers design a framework for communicative games [219], taking Werewolf as a representative case for exploring its feasibility, as shown in Fig. 8. Some emergent strategic behaviors can be observed during gameplay such as trust, confrontation, camouflage, and leadership.

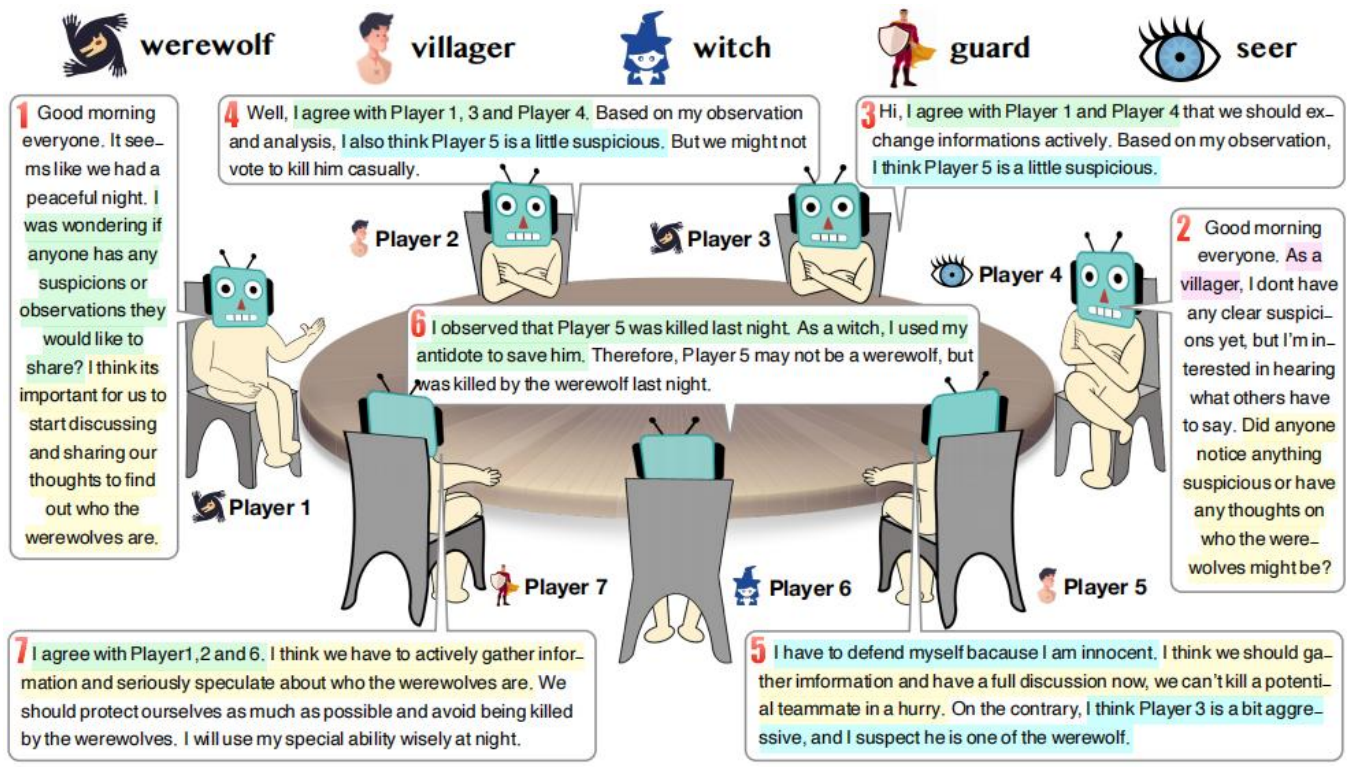

Fig. 8. A snapshot of Werewolf game (Xu et al. [219]).

### 3) Social Simulation

Conducting experiments in the real world is often expensive, anti-ethical, or even impossible. In contrast, agent-based simulation allows researchers to construct hypothetical experimental scenarios that simulate a range of social phenomena, such as the spread of harmful information and social emergence. Researchers are involved in observing and intervening in the system at both the macro and micro levels, enabling them to study counterfactual relationships between events, which can help decision makers to formulate more rational rules or policies. For example, Gao et al. [169] exploited the powerful human-like capabilities of LLM in perception, reasoning and behavior to build a social-network simulation system. Parametrix.ai released GAEA [220], which contains two subsystems, as shown in Fig. 9: (i) Environment system is responsible for interaction and feedback with massive AI NPCs, and drives the self-operation of the entire AI NPC ecosystem; (ii) Soul System uses LLM to help NPCs plan their daily behaviors to create AI NPCs with "life". These two systems interact under the feedback mechanism of GAEA and exert influence on each other, allowing the "AI society" to operate sustainably.

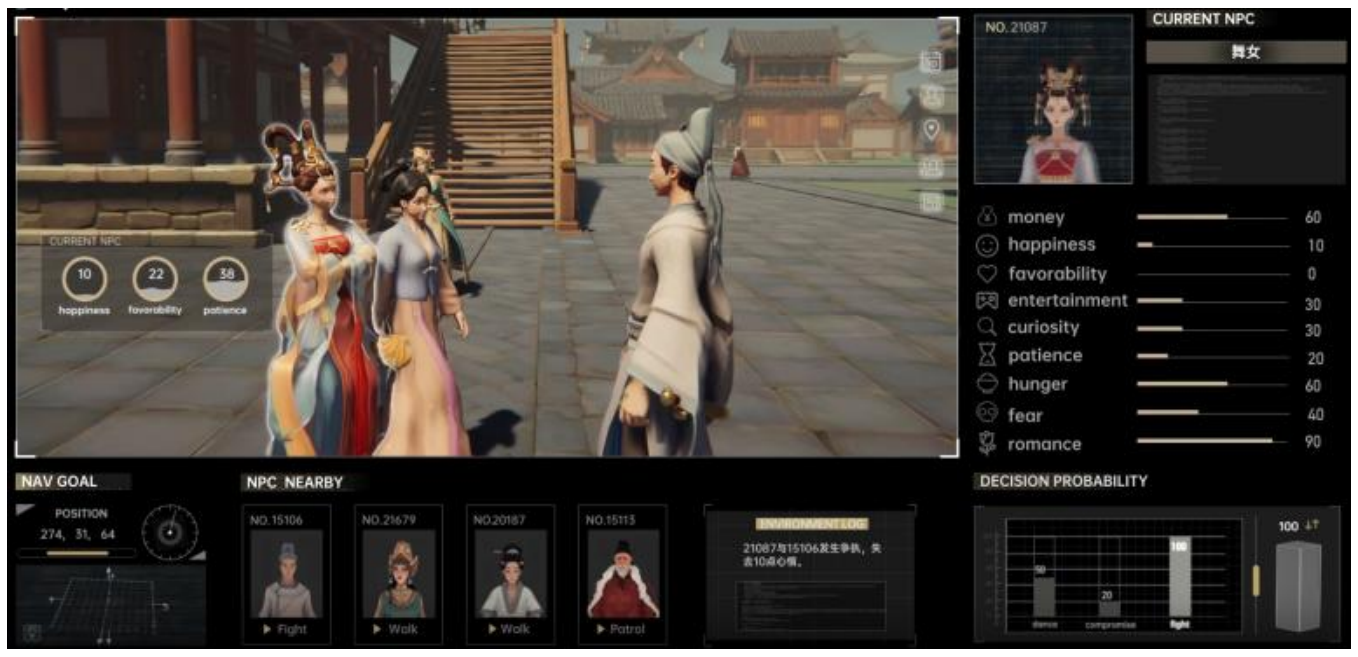

Fig. 9. GAEA: Massive AI-driven NPC System [220].

## IV. Computational Experiments for LLM-based Agent

Computational experiments provide a new way to explain, illustrate, guide and reshape complex phenomena in the real world, which can realize the quantitative analysis of complex systems by algorithmising counterfactual [92]. As shown in Fig. 1, computational experiments offer a distinctive "generative explanation" pathway, which encompasses both generative deduction and generative experiments [221].

### *A. Explainability Enhancement for LLM-based Agent*

LLM has powerful problem-solving capabilities, thanks to its ultra-large model structure and massive learning data. But this also increases model complexity and data dependence of LLM [35], and the property of black-box also follows [36]. In addition, the uncertainty of output results also makes existing objective evaluation indicators ineffective in reflecting the explainability of LLM [37,181]. Therefore, it is very challenging to explore the explainability of LLM from the perspective of model and data. Through a variety of combination experiments, computational experiments can help LLM-based Agent in the artificial society clarify the causal relationships between characteristic factors, individual behaviors, and social emergence, which is positive to the explainability of LLM-based Agent.

Lots of researchers focus on task-centric evaluation of LLM-based agents to figure out the limits of their ability. **(i) Objective Evaluation.** The evaluation tasks include close-book question-answering (QA) based knowledge testing [222,223], multi-turn dialogue [224], reasoning [155,225,226], social evaluation [168,227], benchmarks [177,228,229] and safety assessment [230]. Although objective evaluation enables quantitative assessment of LLM-based agent capabilities, current techniques have limitations in measuring general capabilities. **(ii) Subjective Evaluation.** In some studies [231,271], the human evaluators directly rank or score the generated results of the LLM-based agents based on some specific perspectives; In [168] and [271], human evaluators are always asked to distinguish between agent and human behaviors. Since the LLM-based agent ultimately serves humans, manual evaluation is indispensable. However, cost, efficiency and bias have also become challenges that need to be solved to achieve manual evaluation.

Although the current research works have effectively evaluated the various capabilities of LLM-based Agent and demonstrated ability comparison with the expected general agents and humans, the explainability has still not been fully studied. As shown in Fig. 10, as a complete analysis method for complex systems, computational experiments can provide a general and complete explainability evaluation framework based on "counterfactual" causal analysis for LLM-based Agent, including five steps: customized application scenarios, construction of evaluation experimental systems, design of evaluation experiments, analysis of evaluation experiments, and verification of experimental results. The focus of this framework is to design diverse evaluation experiments to achieve causal analysis based on "counterfactual", thereby revealing the explainability of LLM-based Agent.

#### 1) Customization of Application Scenarios

Computational experiments propose a docking framework to bridge the gap between the diversity of customized application scenarios and the complexity of artificial society. In the framework, the docking specification is based on the basic elements of artificial society (agent model, environment model, and rules model) to realize the extraction of domain knowledge and the integration of artificial society model. Furthermore, it is expected to provide an interactive interface similar to AgentSims [171] where researchers can create agents, construct buildings and equipment in a graphical interface, focusing on the rationality of experiment design, free from complex background driving mechanisms. In the end, the threshold of applying computational experiments method in different fields was greatly reduced.

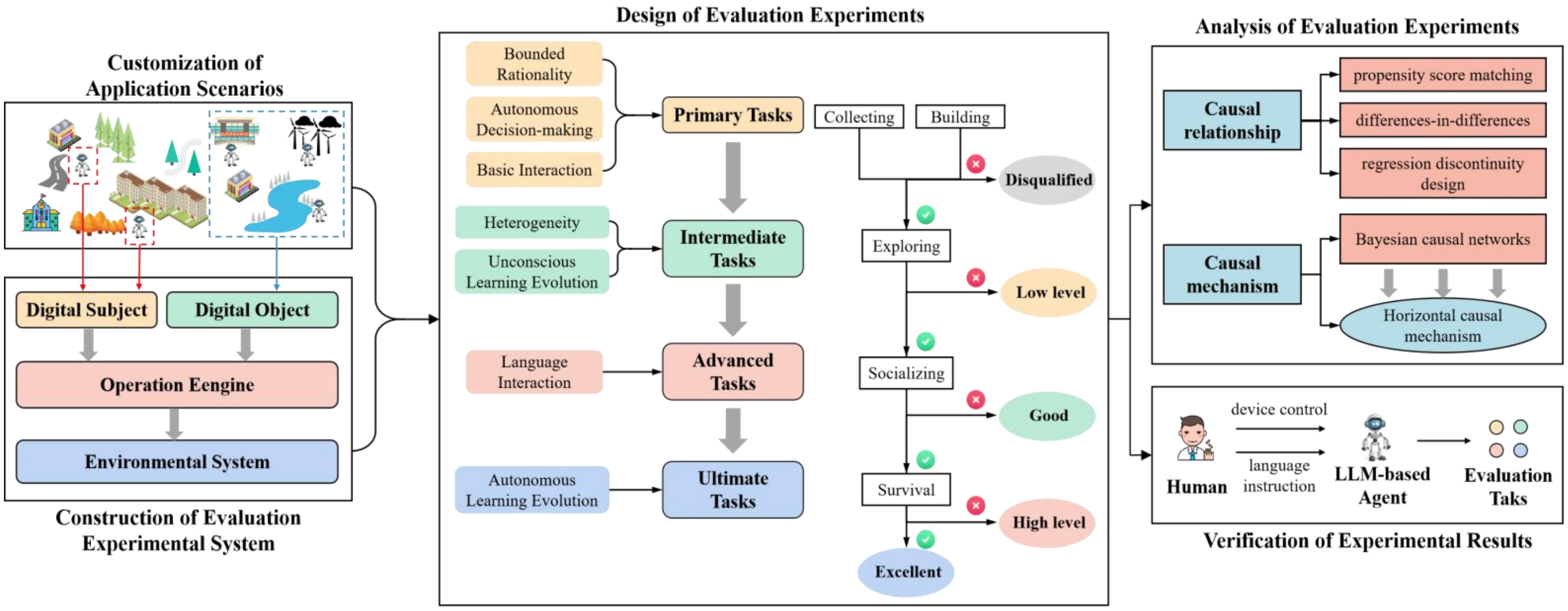

Fig. 10. Explainability evaluation framework for LLM-based Agent.

**2) Construction of Evaluation Experimental System**

Evaluation experimental system consists of four parts [92]: (i) *Digital subject*. By combining different model elements, various LLM-based agents can be generated according to specific demands. (ii) *Digital object*. Environment models can be generated based on specific demands. (iii) *Operation engine*. Different levels of models can be integrated through digital threads to achieve dynamic operation. (iv) *Experimental system*. Various models and technologies are connected in series to construct different computational experimental scenarios.

**3) Design of Evaluation Experiments**

Computational experiments can design a variety of experiments to evaluate the performance of LLM-based Agent in artificial society (Generative Experiments), which can help it to clarify the causal relationship between complex social features and its output behaviors [92], thereby enhancing its explainability.

Generative experiments is used to validate certain theories or hypotheses by introducing artificial interventions into artificial societies and generating data that is highly suitable for investigating causal relationships [232]. As shown in the middle row of Fig. 1, a large number of repeated experiments are carried out in customized application scenarios, and the set of controllable factors in the system that have the greatest impact on the output results can be determined. Furthermore, it is necessary to set up multiple groups of controlled experiments and purposefully change the parameters of customized application scenarios or apply external intervention (counterfactual) to observe the performance of LLM-based agents in the experiments. Here are some types of experiments that can evaluate various practical capabilities of LLM-based Agent:

a) *Collecting.* These tasks are specifically designed to evaluate LLM-based agents' capability in tool use, resource acquisition and environmental awareness. This means that LLM-based agents not only need to identify and aggregate specific resources, but also should navigate through a variety of environments while being aware of their surroundings and the available tools at their disposal [177, 228, 233].

b) *Exploring.* These tasks are designed to be complicated in order to assess LLM-based agents' sense of direction, understanding of foreign environments, and desire to explore. These tasks measure the ability of LLM-based agents to actively understand the world and interact with the various elements [234,235].

c) *Building.* These tasks are devised to evaluate LLM-based agents' aptitude in structural reasoning, spatial organization, and its capability to interact with and manipulate the environment to create specific structures or outcomes. Building requires the interplay of planning, creativity, and understanding of neighbourhood attributes [236,237].

d) *Socializing.* These tasks are used to evaluate whether LLM-based agents can engage in seamless interactions with humans and other agents. These tasks encompass two aspects: (i) the ability to understand literal meanings, implicit meanings and relevant social knowledge (e.g., humour, sarcasm, aggressiveness, and emotion), and to generate appropriate responses [224]; and (ii) the ability to complete the task effectively in an unfamiliar environment through cooperation [215].

e) *Survival.* These tasks aim to analyse the ability of LLM-based agents to ensure their own survival, their proficiency in combat scenarios, and their ability to interact with the environment. Survival requires a complex balance of offensive, defensive and sustaining related actions. These tasks are designed to ensure a comprehensive assessment of these skills for LLM-based agents [166].

**4) Analysis of Evaluation Experiments**

Hofman et al. [238] claimed that causality needs more attention, which can improve the predictive quality of the model. One of the crowning achievements of contemporary work on causation is the formalization of counterfactual reasoning in graphical representations, which is what researchers use to encode scientific knowledge [239]. Each structural equation model determines the "truth value" of each counterfactual sentence. Then an algorithm can determine the probabilities of sentences that can be estimated from experimental or observational studies, or combinations thereof [240,241,242]. In order to study the explainability of LLM-based Agent, it is not only necessary to achieve cause discovery based on experimental data, but also to reveal the mechanism of the cause on the result. [243,244].

a) *Causal relationship.* Researchers usually use some statistical models to analyze the causal relationship between variables and quantitatively evaluate the causal effect of causes on results. Causal relationship here is based on the dependence perspective, which means the dependence and correlation (counterfactual) between causes and results (events, variables, phenomena). Social science researchers commonly adopt potential outcome model to conduct causal analysis: demonstrating a mathematical definition of causality in a counterfactual framework and quantitatively estimating the causal effect of causes on results through statistical models. Common methods include propensity score matching [245], inverse probability weighted estimation [246], regression discontinuity design [247], and differences-in-differences [248]. Furthermore, machine learning can be combined with the above methods to improve their performance. For example, Regression Discontinuity Design can be combined with Gaussian Regression and Bayesian Regression to better fit experimental groups and control groups on either side of the breakpoint [249].

b) *Causal mechanism.* In this evaluation framework, horizontal causal mechanism theory [250] is used to analyze the evaluation experiments. From the perspective of horizontal causal mechanism theory, causal mechanism is interpreted as a network of variables that stand in particularly robust relations, which means the modular sets of entities connected by relations of counterfactual dependence [251]. Therefore, it is necessary to seek the causal structure between multiple variables in a counterfactual framework. A common approach is to learn Bayesian Causal Networks which is a probabilistic graphical model that represents causal relationships and causal structures among multiple variables [252]. The most inspiring work among the current research of social sciences is applying Bayesian Causal Networks to explore relationships among variables in survey data. For example, Craig et al. [253] utilized Bayesian Causal Networks to conduct exploratory analysis on data of Add Health to study the influencing factors of adolescent depression and the relationship between the influencing factors.

**5) Verification of experimental results**

At present, researchers have proposed various validity verification methods for computational experiments [254]. However, this framework is aimed at examining the abilities of LLM-based Agent in customized scenarios to enhance its explainability, which means verification of experimental results is to examine whether the evaluation experiments can represent the corresponding abilities of LLM-based Agent, i.e. **verification of evaluation experiments**. Human is the best criterion to achieve the verification. Turing Test is a highly meaningful and promising approach for assessing intelligent agents to evaluate whether these intelligent systems can represent human-like intelligence [255]. Therefore, a number of human players could be allowed to participate in the same evaluation experiments by controlling agents or providing decision-making instructions to agents. The data generated by LLM-based Agent is compared with that generated by human players to infer whether the evaluation experiments can "appropriately" reflect the abilities of LLM-based Agent.

### *B. Decision Intelligence for applications of LLM-based Agent*

Except for enhancing the explainability of LLM-based Agent, computational experiments can also provide LLM-based Agent with "virtual fact" situation deduction of complex social systems (Generative Deduction), which integrates social cognitive theory, and reveals causal relationships between individual behaviors and complex social phenomena. This can enable LLM-based Agent to dynamically optimize various types of decisions and achieve decision intelligence in complex and uncertain environments [256,257].

Decision intelligence refers to the realization of informed decision-making through real-time and effective perception and analysis of big data, and conducting scenario deductions and situation predictions from a more proactive and comprehensive perspective, meanwhile applying these forward-looking analysis to decision-making. Some studies have revealed the auxiliary decision-making role of decision intelligence in intelligence analysis [258], medical diagnosis [259], business intelligence [260] and other application fields, reflecting the important research value of decision intelligence. As a general problem solver that researchers have high hopes for, LLM-based Agent needs decision intelligence to deal with complex and uncertain social systems, thereby improving its general problem-solving capabilities. However, current work focuses on learning from past knowledge to improve the effectiveness of LLM's decision-making [261], lacking active prediction and assessment for social system situations; and most decisions are modeled in the process [262], which damages maintainability, scalability and flexibility of the process of decision-making.

Computational experiments can build artificial societies in computers and simulate the development trends of real social systems in virtual space to provide managers with a future perspective to achieve decision intelligence. Madhav Marathe and his lab developed a web-based tool that allows public health officials to conduct large-scale epidemiological simulations and hypothetical analyses [263,264]. Users can specify key variables, such as regions of interest (from a single city to the whole country) [265], and diseases of interest (COVID-19) [266]. By using the tool's built-in maps and graphics, users can see the entire simulation realised in front of them, as well as being able to see the likely effects of the model's suggested treatment options.

Therefore, as a method of simulating future scenarios of complex social systems, computational experiments can provide LLM-based Agent with forward-looking analysis of various decisions, that is, a deep understanding of the complex relationship between each decision and its effects; this method also helps LLM-based Agent realize decision-making scheduling management, that is, dynamically optimize the formulation and implementation of various decisions in complex and uncertain environments to achieve expected goals better. The focus of this process is generative deduction, which employs deductive simulation to replicate real-world phenomena, or construct and observe potential alternative worlds.

As shown in Fig. 11, a symbiotic system with dynamic feedback can be established between computational experiments and LLM-based Agent, including knowledge-based decision-making mechanism (Decision), decision deduction based on multiple agents (Generative Deduction), causal inference, and decision mechanism update (Prompting).

**1) Decision**

LLMs are trained on large textual datasets, and can generate a suitable text as a response for any textual input. LLM-based Agent utilizes LLM as its brain and can combine the memory module to make a variety of decisions based on environmental changes and state changes. Subsequently, it can invoke the computational experiments toolkit to deduce the possible social phenomena that may result from these decisions.

**2) Generative Deduction**

Computational experiments build an artificial society to describe the real system where LLM-based agents live, and design experiments to simulate different decisions so that researchers can observe the development trends of the artificial society and arising complex phenomena. As shown in the bottom row of Fig. 1, with the continuous interaction of a vast number of LLM-based agents, the entire artificial society generates new content which influences the future decisions of LLM-based agents, thus achieving a dynamic equilibrium and self-sustaining ecosystem, ultimately resulting in the emergence of various story-lines [267]. It allows for observing and identifying how the causal factors (independent variables) are responsible for the observed changes in the output response (dependent variables). For generative deduction, the future results caused by a certain decision and the way to achieve these future results are both uncertain. The goal of deduction is to predict possible future scenarios so that researchers or LLM-based Agent can optimize existing decisions, i.e. Decision Intelligence [268].

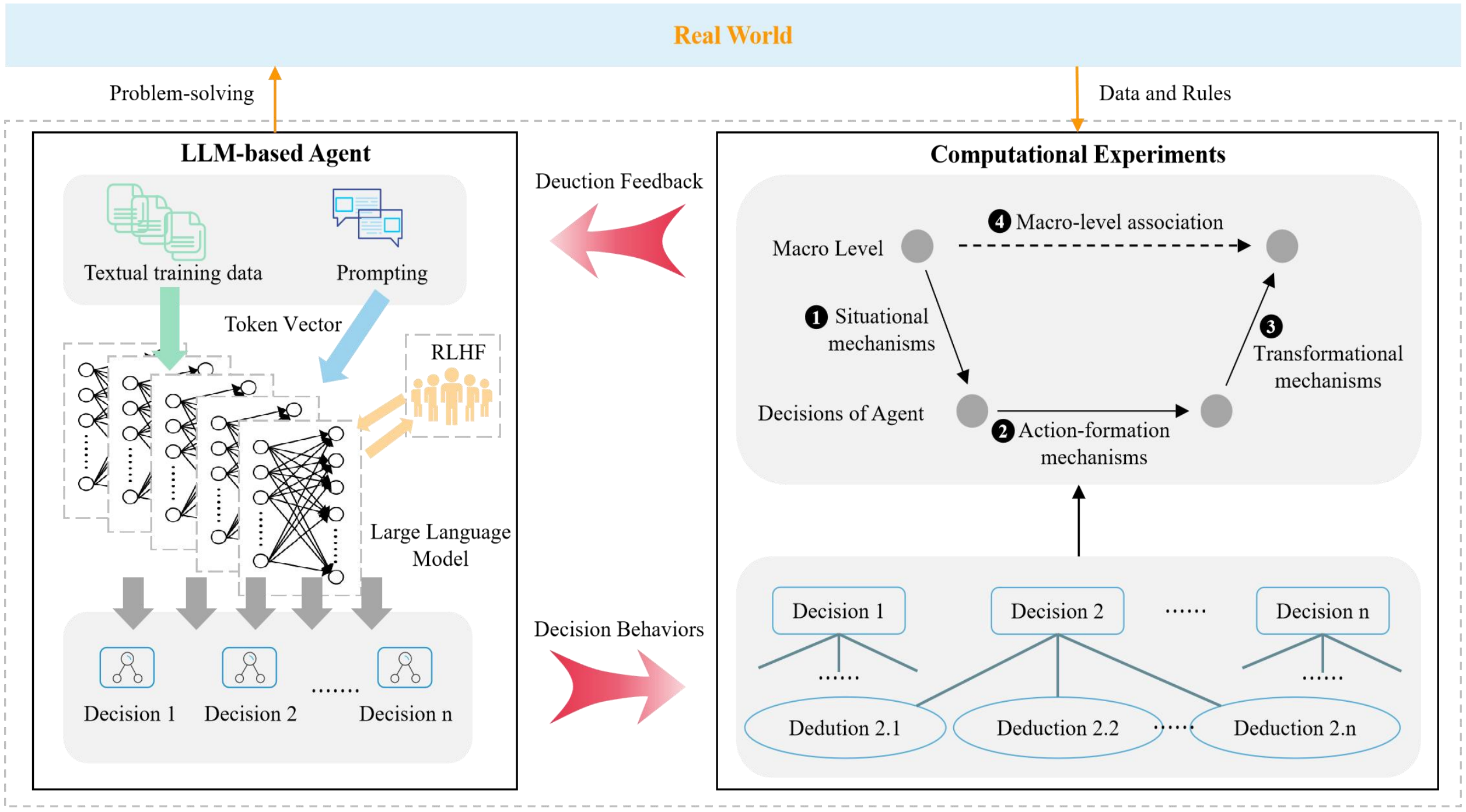

Fig. 11. Decision intelligence of LLM-based Agent with computational experiments.

**3) Causal Inference**

In order to enhance the decision-making mechanism of LLM-based Agent, computational experiments need to conduct causal inference on different simulation results generated by Generative Deduction, so that LLM-based Agent can clarify the causal relationship and causal mechanism between each decision and the complex social phenomena. **(i) Causal relationship** here is based on a productive perspective, which means that the occurrence of one event (or phenomenon) can cause the occurrence of another event (or phenomenon), emphasizing the process by which the cause "leads to" the result. **(ii) Causal mechanism** here is from a generative perspective. "Mechanism" refers to the stable relationship reflected from interactions among individuals in complex systems; mechanistic explanation is to explore how the interactions of individuals emerges macroscopic phenomena in complex systems [269]. Vertical causal mechanism not only focuses on the structure, but also on the process including behaviors and changes: the entity must be in the appropriate position and direction with the appropriate structure; its behaviors must be limited by sequence, speed and time; the mechanism need to show how the entity's behavior at one stage affects its behaviors at subsequent stages [270].

**4) Prompting**

Computational experiments help LLM-based Agent analyze the social phenomena that may be caused by various decisions and the causal relationships between them, and then feed these results back to LLM-based Agent to update the decision-making mechanism. The most commonly used method (In-context learning [194] and CoT [187]) is to convert these causal relationships into natural language instances and provide them as prompts to LLM-based agents to optimizing the decision-making mechanism without changing the original parameters of LLM.

### *C. Application Cases*

LLM-based Agent has strong general problem-solving capabilities and can be used as a decision-making assistant to assist humans in completing task planning. After using computational experiments to achieve decision intelligence, LLM-based Agent can deduce the possible social impacts of various decisions, thereby providing humans with the most reasonable decisions and improving quality and efficiency of human work.

**1) Reasonable Decision**

In the process of dealing with daily affairs, humans usually need to deduce the possible results of various decisions in their brains and choose the most reasonable decision to execute. This requires a lot of physical strength and energy. Therefore, people can ask LLM-based Agent for reasonable decisions by providing the actual situation and even some strategy options [271,272,273]. In [271], the researcher focused on understanding the discourse structure and persuasive elements of political speech with the help of LLM-based Agent. In [272], LLM-based agents were provided with specific characteristics such as talents, preferences, and personality to explore human economic behaviors in simulated scenarios. In [273], LLM-based Agent is used for ideology detection, Fig. 12 shows that there are significant differences between LLMs and traditional media in terms of opinion shaping process, opinion interaction and opinion output. This work suggested that LLMs have a unique and positive effect on the collective opinion difference.

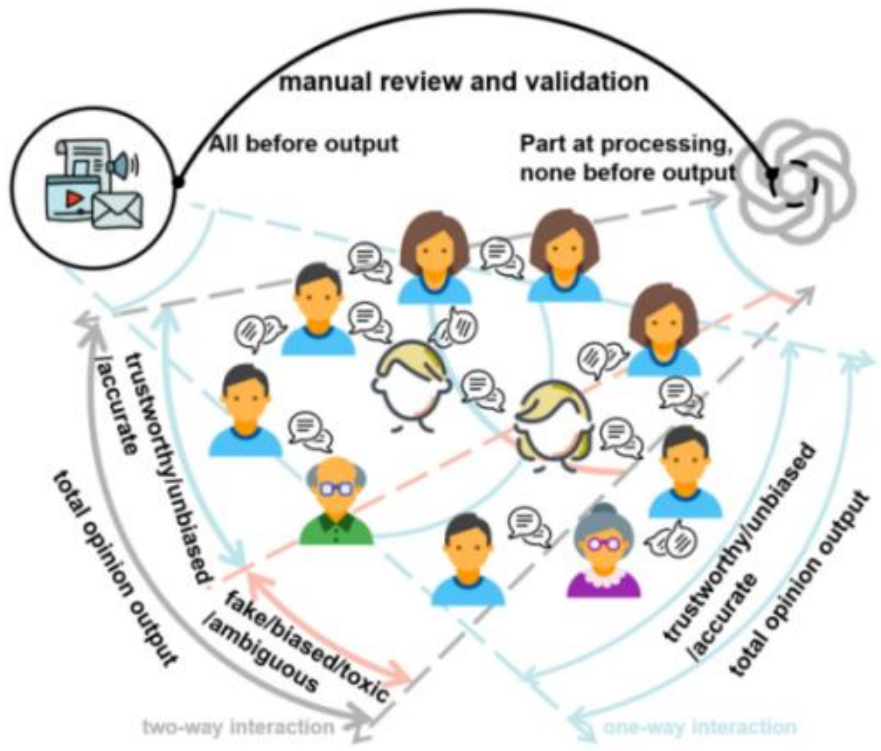

Fig. 12. LLMs and traditional media (Li et al. [273]).

### 2) Improved Quality of Work

LLM-based Agent shows good performance in software development, engineering design, etc., helping humans complete the overall software development process or optimizing structural design, etc., and improving the quality of human work [274,275,276]. [274] proposed an interactive framework for grounded language understanding tasks, which enables humans to interact with AI agents through four different forms of help feedback, to provide high level tips based on concepts relevant for the final task. This work [275] develops LLift, a fully automated framework that interfaces with both a static analysis tool and an LLM. According to LLift, researchers can overcome a number of challenges, including bug-specific modeling, the large problem scope, the non-deterministic nature of LLMs, etc. Dong et al. [276] present a self-collaboration framework for code generation employing LLMs, which assembles an elementary team consisting of three ChatGPT roles (i.e., analyst, coder, and tester) responsible for software development's analysis, coding, and testing stages, as shown in Fig. 13.

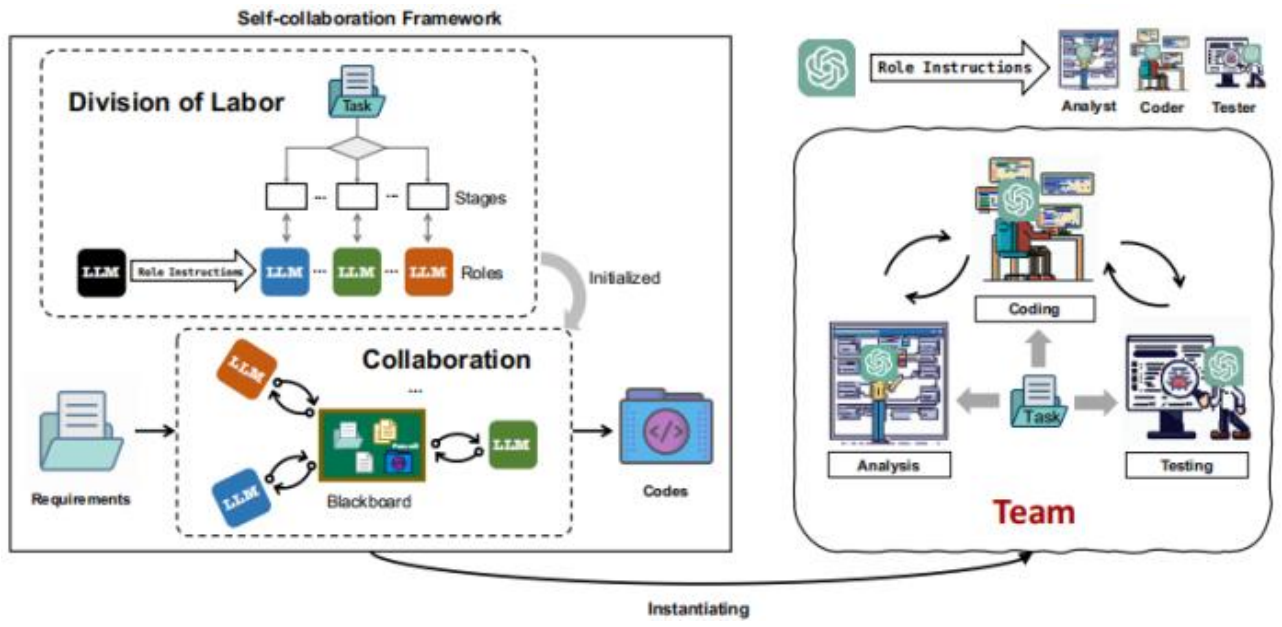

Fig. 13. Self-collaboration for code generation (Dong et al. [276]).

### 3) Improved Efficiency of Work

Various fields in real life contain diverse data, which requires a large number of time and human resources to be collected, organized and synthesized. LLM-based Agent has shown strong capabilities in language understanding and using text-processing tools [277,278]. Boiko et al. [277] also looked into LLM-empowered agents for scientific discovery, which can utilize tools to browse the Internet, read documentation, execute code, call robotic experiment APIs, and leverage other LLMs. ChatMOF [278] extracts key details from textual inputs and delivers appropriate responses, thus eliminating the necessity for rigid structured queries, as shown in Fig. 14.

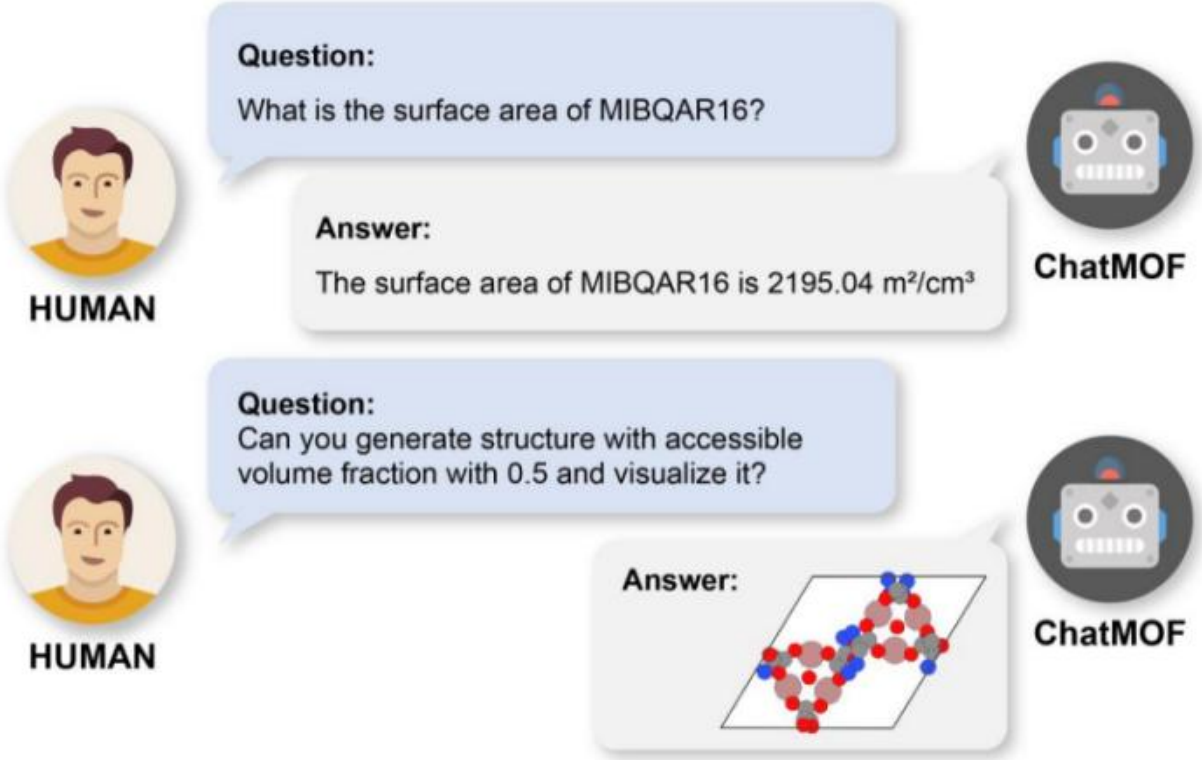

Fig. 14. A conceptual image of ChatMOF (Kang et al. [278]).

## V. Discussion

As a classical method to study complex systems, computational experiments have a mature architecture with complete theoretical system support. Although LLM has not been introduced for a long time, it has been popular all over the world, and its powerful performance in the field of Agent has been favoured by many researchers. However, the combination of computational experiments and LLM-based Agent can never be achieved overnight. Therefore, this chapter discusses the challenges that may be encountered during the combination.

### *A. How to adapt LLM-based Agent to simulation scenarios?*

Recently, LLM-based Agent demonstrates strong general problem solving ability and can build artificial societies that reveal actual laws. However, in the process of simulation, researchers need to focus on whether LLM-based Agent can represent humans in actual scenarios. Then the ensuing question is how to ensure that various attributes of LLM-based Agent are adapted to the simulation scenario.

LLM-based Agent is usually required to play specific roles to perform different tasks in social simulation problems. For example, ChatLaw [279] can simulate a community of lawyers to provide effective legal explanations, similar cases and sound advice to other agents in an artificial society. However, for some occupations with less attention or unfamiliar fields (e.g., crime [280]), LLM-based Agent may not be able to model them well. In addition, the strong generalized problem-solving ability of LLM leads LLM-based Agent to exhibit high intelligence that is incompatible with its simulation context, which in turn leads to hallucinatory problems [38]. Another problem is that LLM-based Agent is not able to model the personality traits of people in each region well, which leads to social simulations that may not reflect real human behaviour well [281].

Potential solutions to these problems include fine-tuning LLMs or well-designed prompts as well as agent's structures. Researchers can collect domain knowledge and human personality data from the target scenario and then use the data to fine-tune LLMs [282,283]. However, how to ensure that the fine-tuned LLMs can still solve general problems efficiently may pose further challenges. In addition to fine-tuning, researchers can also design perfect prompts [284] or improve the structure for LLM-based Agent according to the target domain or target scenario, so that LLM-based Agent can be adapted to the simulated scenario. However, due to the complexity of the target scenario, perfecting effective prompts as well as suitable structures is not easy to achieve.

### *B. How to construct Parallel Society for the combination of LLM-based Agent and Computational Experiments?*

In the past, people needed to pay a high price, even their lives, in the physical world in exchange for experience and wisdom in the intellectual world. Now, human beings can live in the virtual world that have the same functions and attributes as the real world, to gain experience for their behaviour in the real world and to enhance their wisdom. The virtual world is parallel society.

The key elements of a parallel society are humans, devices and things, which depend on each other and develop in harmony. The fusion of these three types of information is the key to enhancing intelligence of machine and realizing the interactive fusion of virtual space and real space [285,286]. Humans provide uncertainty, diversity and complexity to agents in the parallel society, and agents internalize these attributes into agility, the ability to focus towards tasks and the ability to converge towards goals [287]. Computational experiments need to achieve parallel optimization, i.e., obtaining real-time and dynamic data from reality, and achieving synchronous calibration between the artificial society and the real society [288]. The artificial society constructed by LLM-based Agent aims at the parallel society, and the following two types of model need to be considered.

#### 1) World Model

Most researchers worked on integrating various deep learning models to achieve AGI before LLM, but because the hierarchy of cognitive models is not deep enough, there are no very significant progress at present. Yann LeCun has also commented on LLM when thinking about AGI and has argued that world model [289] is critical to human perception of the world. Regardless of whether LLM includes the cognition of the world or not, agent must have an accurate understanding of the world in order to make correct decisions. Therefore, before constructing an artificial society, LLM-based Agent must have accurate knowledge of the simulation scenario, including social structure, social relationship network, operation mechanism, etc.

#### 2) Rule Model

Artificial societies constructed by LLM-based Agent must encompass the full range of complex characteristics of complex systems [94,95]: **(i) Autonomy**, where agents make decisions based on their perceptions, memories and behaviour sets; **(ii) Evolution**, where the artificial society conforms to the real-world rules of "survival of the fittest"; **(iii) Interaction**, where interaction rules are designed for agents, including cooperation, games, competition, topology, etc.; **(iv) Emergence**, where interactions among agents drive the emergence of the society.

### *C. How to implement invoking computational experiments for LLM-based Agent?*

Currently, computational experiments have a well-defined methodological framework and can be invoked by LLM-based agents as a deduction toolkit to simulate and deduce the scenarios they are in [290,291]. The method can assess the possible impact of individual decisions, and help LLM-based Agent optimize decisions with a future perspective. However, there are some challenges for LLM-based agents to invoke computational experiments. Firstly, LLM-based Agent needs to understand the state of its environment and the impact of its own behaviour on the state of the environment. This is the basis on which computational experiments can build an artificial society for LLM-based Agent [292]. Secondly, as an inference toolkit, computational experiments require LLM-based Agent to specify the influencing factors, experimental conditions, and value-taking strategies related to the experimental design, so as to construct the desired hypothetical environment and complete the deduction. Finally, LLM-based Agent needs to have the ability to understand the results of computational experiments and have a clear understanding of the causal relationship between behaviors and phenomena.

Potential solutions to these challenges can be realized from two perspectives: (i) LLM-based Agent can complete computational experiments step by step like humans. Researchers can use the case of computational experiments to fine-tune LLMs so that LLM-based Agent understands each step of the computational experiments and progressively completes simulation and deduction. (ii) Automated Computational Experiments Toolkit. The entire workflow is a collaboration of multiple LLM-based agents responsible for each of the five steps of computational experiments, similar to the currently implemented collaborative software development process [155]. LLM-based Agent uses natural language to describe the task scenario and the decision set. The computational experiments toolkit receives the language information and opens the workflow for automated deduction. Upon completion of the experiments, the toolkit presents the results in natural language to the LLM-based Agent to optimize decision-making.

## VI. Conclusion

The rapid development of new generation of information communication technology has profoundly changed the relationship between people, environment, and society. The resulting CPSS presents complex features such as highly open, dynamic and interactive. As a classic method for the research of complex system, computational experiments need to be combined with LLM-based Agent to complement each other. This can not only promote the artificial society to perfectly reflect the real system and enable computational experiments to achieve breakthroughs in the research of complex system, but also enhance the explainability of LLM-based Agent through generative experiments and help LLM-based Agent achieve decision intelligence by generative deduction, thus providing a new direction on the road to realizing Artificial General Intelligence. This paper aims to introduce LLM-based Agent to Computational Experiments. Therefore, it's necessary to review the traditional methods of constructing agents and artificial societies in computational experiments, which can help researchers clarify the current challenges encountered in computational experiments. After that, this paper compares LLM-based Agent with agents built by traditional methods in computational experiments, showing the enhancements that LLM-based Agent brings to Agent as well as to Artificial Society. Furthermore, this paper describes how computational experiments can improve the explainability of LLM-based Agent and enhance its ability to aid decision-making from a perspective of "generative explanation". The advantages that they can provide to each other are discussed from the two perspectives: LLM-based Agent for CE and CE for LLM-based Agent. Finally, this paper discusses some challenges and the future trends in realizing the combination of computational experiments and LLM-based Agents and proposes reasonable solutions to provide corresponding guidance for subsequent research work.

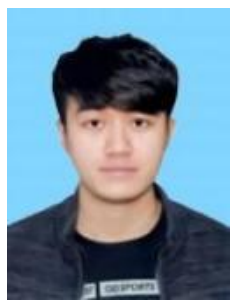

**Qun Ma** is currently pursuing the Ph.D. degree in the College of Intelligence and Computing, Tianjin University, Tianjin, China.

His research interest covers computational experiments and agent modeling based on large language model.

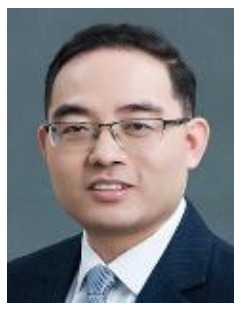

**Xiao Xue** received the Ph.D. degree in control theory and control engineering from the Institute of Automation, Chinese Academy of Sciences, Beijing, China, in 2007.

He is currently a Professor with the School of Computer Software, College of Intelligence and Computing, Tianjin University, Tianjin, China. His main research interests include service computing, computational experiment, and social computing.

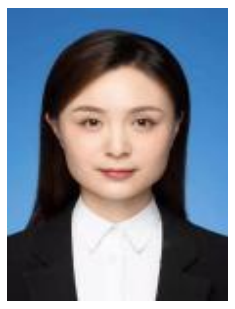

**De-Yu Zhou** is currently pursuing the Ph.D. degree in artificial intelligence with the School of Software, Shandong University, Jinan, China.

Her research interest covers service computing and computational experiments.

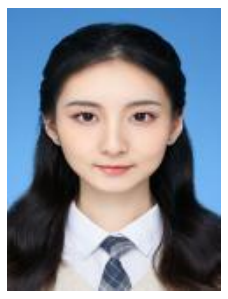

**Xiang-Ning Yu** is the Ph.D. candidate student at the College of Intelligence and Computing, Tianjin University, Tianjin, China.

Her research interest covers service computing and computational experiments.

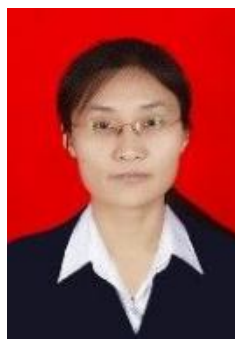

**Dong-Hua Liu** received the Ph.D. degree from Wuhan University, Wuhan, China, in 2021.

She is currently an associate researcher at the China Waterborne Transport Research Institute. Her research interests include data mining and intelligent recommendation.

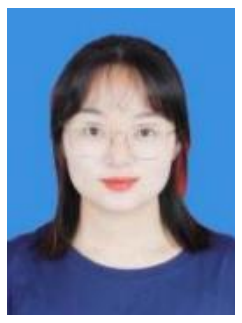

**Xu-Wen Zhang** is the Ph.D. candidate student at the College of Intelligence and Computing, Tianjin University, Tianjin, China.

Her research interest covers service computing and computational experiments.

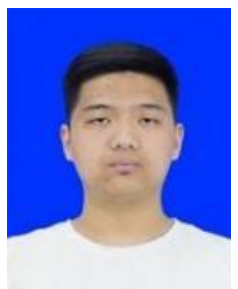

**Zi-Han Zhao** is the graduate student with the College of Computer and Information, Hohai University.

His current research interests include computational experiment.

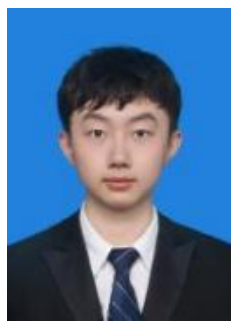

**Yi-Fan Shen** is currently pursuing the master’s degree with the College of Intelligence and Computing, Tianjin University, Tianjin, China.

His research interest covers service computing, computational experiments, swarm intelligence and large language model based agents.

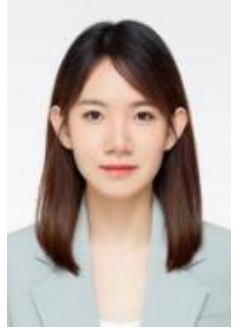

**Pei-Lin Ji** is currently pursuing the master’s degree in electronic information with the college of intelligence and computing, Tianjin University, Tianjin, China.

Her research interest covers service computing and computational experiments.

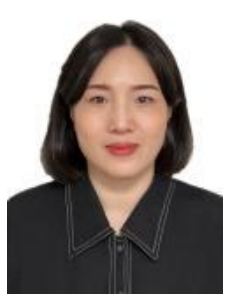

**Juan-Juan Li** received her Ph.D. degree in control science and engineering from Beijing Institute of Technology, Beijing, China, in 2021.

She is currently an associate professor with the State Key Laboratory of Multimodal Artificial Intelligence Systems, Institute of Automation, Chinese Academy of Sciences, Beijing, China. Her research interests include blockchain, DAO, and decision intelligence.

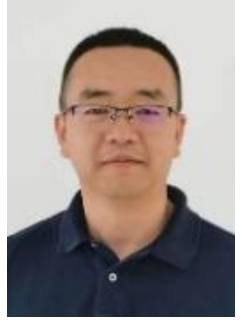

**Gang Wang** is currently a professor at Information School, Xi’an University of Financial and Economics. Postdoctoral Fellow for University of Texas at San Antonio.

His current research interests include Trust Management, Privacy Security and Social Computing, Cloud Computing and AIoT.

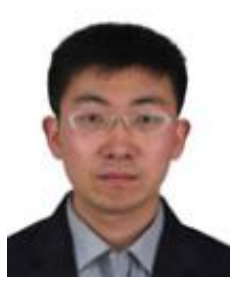

**Wan-Peng Ma** is an engineer in Army Aviation Research Institute. He received the Ph.D. degree in materials scienee and engineering from Academy of Armored Forces Engineering.

His research interests include multi-agent systems, design of experiments, and combat modeling.